\documentclass[preprint,12pt]{elsarticle}

\usepackage{amssymb}

\usepackage[a4paper, top=2.54cm, bottom=2.54cm, left=3.18cm, right=3.18cm]{geometry}
\usepackage{enumitem} 

\usepackage{adjustbox}

\usepackage{caption}
\usepackage{subcaption}
\usepackage{multirow}
\usepackage{booktabs}
\usepackage{array}
\usepackage{textcomp}

\usepackage[breaklinks,colorlinks=false,linkcolor=black,citecolor=black,urlcolor=black]{hyperref}
\usepackage[hyphenbreaks]{breakurl}

\usepackage{CJKutf8}

\journal{Information Fusion}

\begin{document}
\begin{CJK}{UTF8}{gbsn}

\begin{frontmatter}



\title{PlaneSAM: Multimodal Plane Instance Segmentation Using the Segment Anything Model}


\author[inst1]{Zhongchen Deng\fnref{equal}}
\ead{dzc@stu.hubu.edu.cn}

\affiliation[inst1]{organization={School of Computer Science and Information Engineering, Hubei University},
            city={Wuhan},
            postcode={430062}, 
            state={Hubei},
            country={China}}

\author[inst2,inst3]{Zhechen Yang\fnref{equal}}
\ead{zc.yang@stu.hubu.edu.cn}

\affiliation[inst2]{organization={School of Artificial Intelligence, Hubei University},
            city={Wuhan},
            postcode={430062}, 
            state={Hubei},
            country={China}}
            
\affiliation[inst3]{organization={Key Laboratory of Intelligent Sensing System and Security (Hubei University), Ministry of Education},
            city={Wuhan},
            postcode={430062}, 
            state={Hubei},
            country={China}}          
            
\author[inst4,inst5,inst6]{Chi Chen\fnref{equal}}

\affiliation[inst4]{organization={State Key Laboratory of Information Engineering in Surveying, Mapping and Remote Sensing, Wuhan University},
            city={Wuhan},
            postcode={430072}, 
            state={Hubei},
            country={China}}
\affiliation[inst5]{organization={Engineering Research Center of Space-Time Data Capturing and Smart Application, the Ministry of Education of P.R.C.},
            city={Wuhan},
            postcode={430072}, 
            state={Hubei},
            country={China}}
\affiliation[inst6]{organization={Institute of Geospatial Intelligence, Wuhan University},
            city={Wuhan},
            postcode={430072}, 
            state={Hubei},
            country={China}}

\author[inst2,inst3]{Cheng Zeng}
\author[inst2,inst3]{Yan Meng\corref{cor1}}
\ead{mengyan@hubu.edu.cn}
\author[inst4,inst5,inst6]{Bisheng Yang\corref{cor1}}
\cortext[cor1]{Corresponding author}
\ead{bshyang@whu.edu.cn}
\fntext[equal]{These authors contributed equally to this work.}

%
%
%

\begin{abstract}
Plane instance segmentation from RGB-D data is a crucial research topic for many downstream tasks, such as indoor 3D reconstruction. However, most existing deep-learning-based methods utilize only information within the RGB bands, neglecting the important role of the depth band in plane instance segmentation. Based on EfficientSAM, a fast version of the Segment Anything Model (SAM), we propose a plane instance segmentation network called PlaneSAM, which can fully integrate the information of the RGB bands (spectral bands) and the D band (geometric band), thereby improving the effectiveness of plane instance segmentation in a multimodal manner. Specifically, we use a dual-complexity backbone, with primarily the simpler branch learning D-band features and primarily the more complex branch learning RGB-band features. Consequently, the backbone can effectively learn D-band feature representations even when D-band training data is limited in scale, retain the powerful RGB-band feature representations of EfficientSAM, and allow the original backbone branch to be fine-tuned for the current task.
To enhance the adaptability of our PlaneSAM to the RGB-D domain, we pretrain our dual-complexity backbone using the segment anything task on large-scale RGB-D data through a self-supervised pretraining strategy based on imperfect pseudo-labels. 
To support the segmentation of large planes, we optimize the loss function combination ratio of EfficientSAM. 
In addition, Faster R-CNN is used as a plane detector, and its predicted bounding boxes are fed into our dual-complexity network as prompts, thereby enabling fully automatic plane instance segmentation. Experimental results show that the proposed PlaneSAM sets a new state-of-the-art (SOTA) performance on the ScanNet dataset, and outperforms previous SOTA approaches in zero-shot transfer on the 2D-3D-S, Matterport3D, and ICL-NUIM RGB-D datasets, while only incurring a 10\% increase in computational overhead compared to EfficientSAM. Our code and trained model will be released publicly.
\end{abstract}




\end{frontmatter}


\section{Introduction}
\label{sec:Introduction}
Plane instance segmentation aims to identify and extract planes from point clouds or image data corresponding to 3D scenes~\cite{Li_ISPRS2024, Tan_ICCV2021, Yan_ISPRS2014, Zhang_PR2022}. 
It has been widely used in many subfields of photogrammetry and computer vision, such as indoor 3D reconstruction~\cite{Fang_ISPRS2021, Li_ISPRS2021, Stotko_ISPRS2019}, indoor mobile mapping~\cite{Xiang_JAG2021}, outdoor 3D reconstruction~\cite{Li_ISPRS2022, Li_RS2020, LiShan_ISPRS2022}, autonomous driving~\cite{Qian_TIV2023}, augmented reality~\cite{Xie_ICRA2021}, and SLAM~\cite{Li_TGRS2023}. 
By segmenting planar regions in 3D scenes, we can obtain important geometric and semantic information for downstream tasks, helping people better understand and process complex visual data. 
Depth data can provide complementary information to spectral data, and the organic combination of the two modalities often yields better results than the use of only single-modal data~\cite{Zhou_IF2023, MM_IF2021}. 
Furthermore, planes are a type of geometric primitive, so incorporating relevant geometric information of the D band is expected to enhance the robustness of plane instance segmentation. 
Therefore, the present study investigates the problem of plane instance segmentation from RGB-D data, with the aim of achieving a higher plane instance segmentation accuracy in a multimodal manner.

For plane instance segmentation, although many state-of-the-art (SOTA) algorithms focus on RGB-D datasets, most of them utilize only the RGB bands (spectral data), overlooking the significant role of the D band (geometric data). 
For example, PlaneRCNN~\cite{Liu_CVPR2019}, PlaneAE~\cite{Yu_CVPR2019}, PlaneSeg~\cite{Zhang_TNNLS2024}, and BT3DPR~\cite{Ren_GM2024} use the RGB-D datasets for deep learning model training and testing, but they only utilize the RGB bands. 
Recognizing that planes are geometric structures, some researchers employ other geometric structures to aid in recovering plane formations. 
For example, the SOTA algorithms PlaneTR~\cite{Tan_ICCV2021} and PlaneAC~\cite{Zhang_PR2024} first extract line segments from RGB bands as geometric clues to assist in plane instance segmentation. 
However, the extraction of geometric clues from RGB bands also relies on spectral features, making these clues susceptible to variations in shooting conditions. 
X-PDNet~\cite{Cao_BMVC2023} utilizes information from the depth estimated based on the RGB bands, making it a sequential method that is prone to errors. 
Furthermore, essentially, it also relies entirely upon the information of the RGB bands.

\begin{figure}[htbp!]
    \centering

    \begin{minipage}{0.05\linewidth}
        \centering
        \adjustbox{valign=c}{\rotatebox{90}{\fontsize{10pt}{10pt}\selectfont{RGB}}}
    \end{minipage}%
    \hspace{-0.3em} 
    \begin{minipage}{0.23\linewidth}
        \centering
        \includegraphics[width=\linewidth]{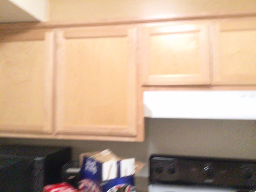}
    \end{minipage}%
    \hspace{0.05em} 
    \begin{minipage}{0.23\linewidth}
        \centering
        \includegraphics[width=\linewidth]{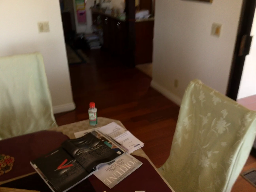}
    \end{minipage}%
    \hspace{0.05em} 
    \begin{minipage}{0.23\linewidth}
        \centering
        \includegraphics[width=\linewidth]{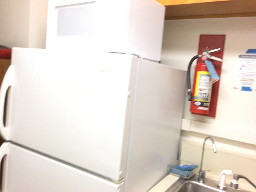}
    \end{minipage}%
    \hspace{0.05em}
    \begin{minipage}{0.23\linewidth}
        \centering
        \includegraphics[width=\linewidth]{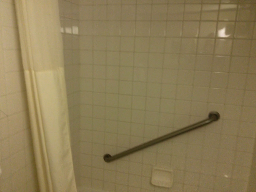}
    \end{minipage}

    \vspace{0.3em} 

    \begin{minipage}{0.05\linewidth}
        \centering
        \adjustbox{valign=c}{\rotatebox{90}{\fontsize{10pt}{10pt}\selectfont{Depth}}}
    \end{minipage}%
    \hspace{-0.3em} 
    \begin{minipage}{0.23\linewidth}
        \centering
        \includegraphics[width=\linewidth]{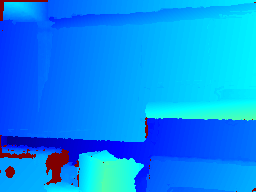}
    \end{minipage}%
    \hspace{0.05em} 
    \begin{minipage}{0.23\linewidth}
        \centering
        \includegraphics[width=\linewidth]{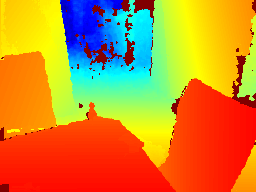}
    \end{minipage}%
    \hspace{0.05em} 
    \begin{minipage}{0.23\linewidth}
        \centering
        \includegraphics[width=\linewidth]{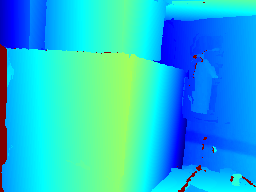}
    \end{minipage}%
    \hspace{0.05em} 
    \begin{minipage}{0.23\linewidth}
        \centering
        \includegraphics[width=\linewidth]{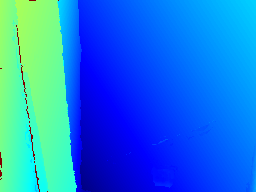}
    \end{minipage}

    \vspace{0.3em} 

    \begin{minipage}{0.05\linewidth}
        \centering
        \adjustbox{valign=c}{\rotatebox{90}{\fontsize{10pt}{10pt}\selectfont{Ground-truth}}}
    \end{minipage}%
    \hspace{-0.3em} 
    \begin{minipage}{0.23\linewidth}
        \centering
        \includegraphics[width=\linewidth]{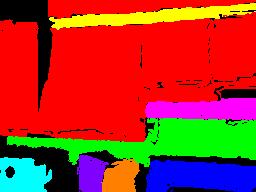}
    \end{minipage}%
    \hspace{0.05em} 
    \begin{minipage}{0.23\linewidth}
        \centering
        \includegraphics[width=\linewidth]{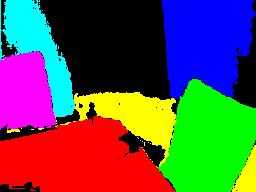}
    \end{minipage}%
    \hspace{0.05em} 
    \begin{minipage}{0.23\linewidth}
        \centering
        \includegraphics[width=\linewidth]{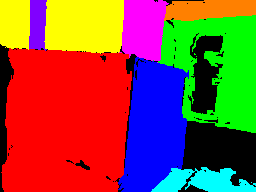}
    \end{minipage}%
    \hspace{0.05em} 
    \begin{minipage}{0.23\linewidth}
        \centering
        \includegraphics[width=\linewidth]{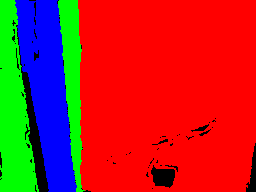}
    \end{minipage}

    \vspace{0.3em} 
 
    \begin{minipage}{0.05\linewidth}
        \centering
        \adjustbox{valign=c}{\rotatebox{90}{\fontsize{10pt}{10pt}\selectfont{PlaneTR}}}
    \end{minipage}%
    \hspace{-0.3em} 
    \begin{minipage}{0.23\linewidth}
        \centering
        \includegraphics[width=\linewidth]{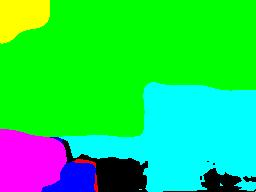}
    \end{minipage}%
    \hspace{0.05em} 
    \begin{minipage}{0.23\linewidth}
        \centering
        \includegraphics[width=\linewidth]{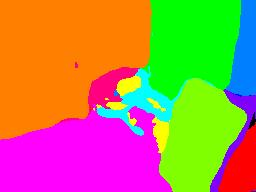}
    \end{minipage}%
    \hspace{0.05em} 
    \begin{minipage}{0.23\linewidth}
        \centering
        \includegraphics[width=\linewidth]{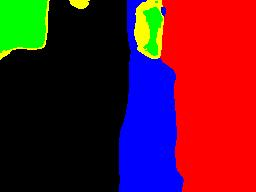}
    \end{minipage}%
    \hspace{0.05em} 
    \begin{minipage}{0.23\linewidth}
        \centering
        \includegraphics[width=\linewidth]{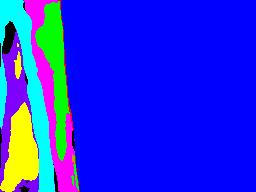}
    \end{minipage}

    \vspace{0.3em} 

    \begin{minipage}{0.05\linewidth}
        \centering
        \adjustbox{valign=c}{\rotatebox{90}{\fontsize{10pt}{10pt}\selectfont{PlaneSAM}}}
    \end{minipage}%
    \hspace{-0.3em} 
    \begin{minipage}{0.23\linewidth}
        \centering
        \includegraphics[width=\linewidth]{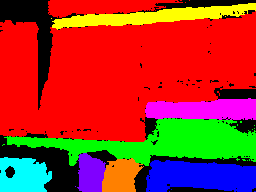}
    \end{minipage}%
    \hspace{0.05em} 
    \begin{minipage}{0.23\linewidth}
        \centering
        \includegraphics[width=\linewidth]{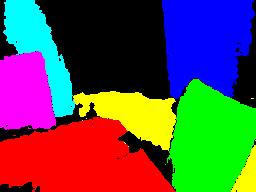}
    \end{minipage}%
    \hspace{0.05em} 
    \begin{minipage}{0.23\linewidth}
        \centering
        \includegraphics[width=\linewidth]{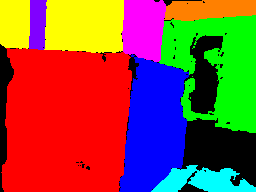}
    \end{minipage}%
    \hspace{0.05em} 
    \begin{minipage}{0.23\linewidth}
        \centering
        \includegraphics[width=\linewidth]{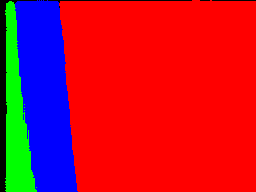}
    \end{minipage}

    \vspace{0.4em} 
    \captionsetup{font=small}  
    \caption{Failure cases of plane segmentation based solely on RGB spectral bands, taking the SOTA algorithm PlaneTR as an example. Each row from top to bottom shows RGB images, depth images, ground-truths, segmentation results of the PlaneTR algorithm, and segmentation results of our PlaneSAM. Areas marked in black in the segmentation results represent non-planar regions.}
    \label{fig:RGBFailures}
\end{figure}

Existing SOTA algorithms achieve acceptable results based solely on the RGB bands (spectral bands) because the spectral variations of typical planar objects in indoor scenes are generally limited. Additionally, although existing RGB-D plane instance segmentation datasets are not very large, they have reached a preliminary scale. However, the segmentation of plane instances from a single-view RGB image is an ill-posed problem. 
Spectral features, in fact, lack robustness as descriptors of plane instances and cannot effectively capture changes in three dimensions, resulting in limited performance for SOTA algorithms that rely solely on RGB spectral features (see Figure~\ref{fig:RGBFailures}). 
These challenges can be handled by appropriate use of all RGB-D bands. 
Therefore, when the data has a D band, we should comprehensively utilize geometric information found on the D band in addition to that found on the spectral bands, instead of focusing on solely the spectral features in the RGB bands and certain geometric clues extracted from those spectral features.

However, training a network model to achieve strongly generalizable plane segmentation performance on RGB-D data is not a trivial task. 
Prior studies on deep learning suggest that extensive, finely annotated training data is typically required for a network to achieve strong generalizability~\cite{SAM2023, CLIP2021, Zhang_RS2024, Devlin_2019Bert, Ao_TGRS2023}. 
However, existing RGB-D training datasets for plane instance segmentation, such as the ScanNet plane segmentation dataset~\cite{Liu_CVPR2018}, are not very large in scale. Additionally, the cost of annotating new large-scale RGB-D datasets for plane instance segmentation is prohibitively high.

\begin{figure}[t!]
    \centering

    \begin{minipage}{0.05\linewidth}
        \centering
        \adjustbox{valign=c}{\rotatebox{90}{\fontsize{10pt}{10pt}\selectfont{RGB}}}
    \end{minipage}%
    \hspace{-0.3em} 
    \begin{minipage}{0.23\linewidth}
        \centering
        \includegraphics[width=\linewidth]{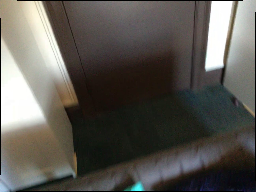}
    \end{minipage}%
    \hspace{0.05em} 
    \begin{minipage}{0.23\linewidth}
        \centering
        \includegraphics[width=\linewidth]{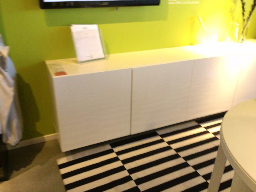}
    \end{minipage}%
    \hspace{0.05em} 
    \begin{minipage}{0.23\linewidth}
        \centering
        \includegraphics[width=\linewidth]{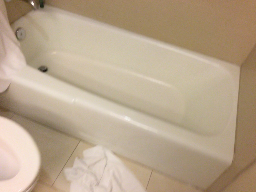}
    \end{minipage}%
    \hspace{0.05em}
    \begin{minipage}{0.23\linewidth}
        \centering
        \includegraphics[width=\linewidth]{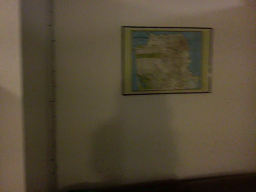}
    \end{minipage}

    \vspace{0.3em} 

    \begin{minipage}{0.05\linewidth}
        \centering
        \adjustbox{valign=c}{\rotatebox{90}{\fontsize{10pt}{10pt}\selectfont{Depth}}}
    \end{minipage}%
    \hspace{-0.3em} 
    \begin{minipage}{0.23\linewidth}
        \centering
        \includegraphics[width=\linewidth]{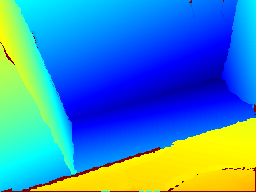}
    \end{minipage}%
    \hspace{0.05em} 
    \begin{minipage}{0.23\linewidth}
        \centering
        \includegraphics[width=\linewidth]{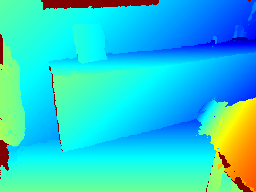}
    \end{minipage}%
    \hspace{0.05em} 
    \begin{minipage}{0.23\linewidth}
        \centering
        \includegraphics[width=\linewidth]{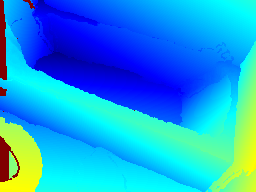}
    \end{minipage}%
    \hspace{0.05em} 
    \begin{minipage}{0.23\linewidth}
        \centering
        \includegraphics[width=\linewidth]{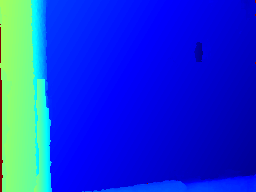}
    \end{minipage}

    \vspace{0.3em} 

    \begin{minipage}{0.05\linewidth}
        \centering
        \adjustbox{valign=c}{\rotatebox{90}{\fontsize{10pt}{10pt}\selectfont{Ground-truth}}}
    \end{minipage}%
    \hspace{-0.3em} 
    \begin{minipage}{0.23\linewidth}
        \centering
        \includegraphics[width=\linewidth]{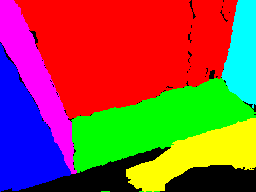}
    \end{minipage}%
    \hspace{0.05em} 
    \begin{minipage}{0.23\linewidth}
        \centering
        \includegraphics[width=\linewidth]{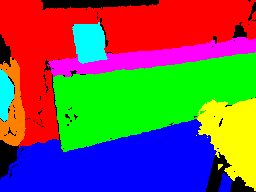}
    \end{minipage}%
    \hspace{0.05em} 
    \begin{minipage}{0.23\linewidth}
        \centering
        \includegraphics[width=\linewidth]{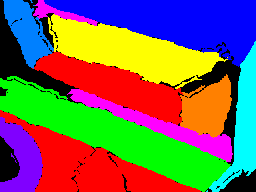}
    \end{minipage}%
    \hspace{0.05em} 
    \begin{minipage}{0.23\linewidth}
        \centering
        \includegraphics[width=\linewidth]{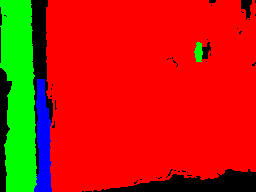}
    \end{minipage}

    \vspace{0.3em} 
 
    \begin{minipage}{0.05\linewidth}
        \centering
        \adjustbox{valign=c}{\rotatebox{90}{\fontsize{10pt}{10pt}\selectfont{PlaneTR}}}
    \end{minipage}%
    \hspace{-0.3em} 
    \begin{minipage}{0.23\linewidth}
        \centering
        \includegraphics[width=\linewidth]{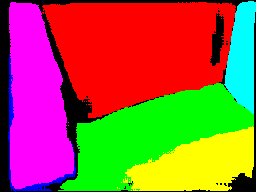}
    \end{minipage}%
    \hspace{0.05em} 
    \begin{minipage}{0.23\linewidth}
        \centering
        \includegraphics[width=\linewidth]{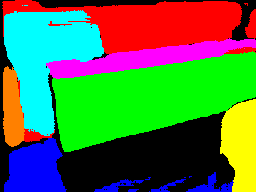}
    \end{minipage}%
    \hspace{0.05em} 
    \begin{minipage}{0.23\linewidth}
        \centering
        \includegraphics[width=\linewidth]{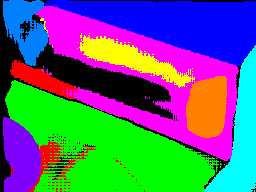}
    \end{minipage}%
    \hspace{0.05em} 
    \begin{minipage}{0.23\linewidth}
        \centering
        \includegraphics[width=\linewidth]{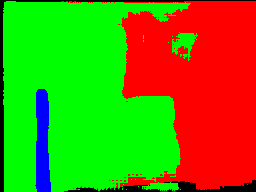}
    \end{minipage}

    \vspace{0.3em} 

    \begin{minipage}{0.05\linewidth}
        \centering
        \adjustbox{valign=c}{\rotatebox{90}{\fontsize{10pt}{10pt}\selectfont{PlaneSAM}}}
    \end{minipage}%
    \hspace{-0.3em} 
    \begin{minipage}{0.23\linewidth}
        \centering
        \includegraphics[width=\linewidth]{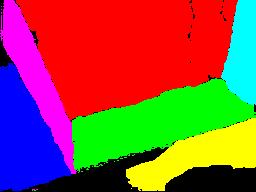}
    \end{minipage}%
    \hspace{0.05em} 
    \begin{minipage}{0.23\linewidth}
        \centering
        \includegraphics[width=\linewidth]{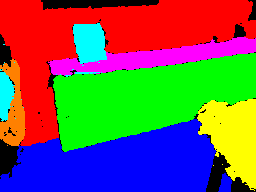}
    \end{minipage}%
    \hspace{0.05em} 
    \begin{minipage}{0.23\linewidth}
        \centering
        \includegraphics[width=\linewidth]{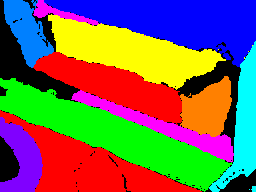}
    \end{minipage}%
    \hspace{0.05em} 
    \begin{minipage}{0.23\linewidth}
        \centering
        \includegraphics[width=\linewidth]{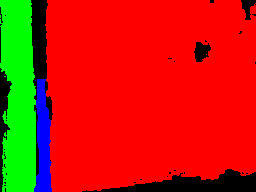}
    \end{minipage}

    \vspace{0.4em} 
    \captionsetup{font=small}  
    \caption{Failure cases of plane segmentation by only fine-tuning the original EfficientSAM backbone. Each row from top to bottom shows RGB images, depth images, ground-truths, results obtained by only fine-tuning the original EfficientSAM backbone, and results obtained by our PlaneSAM. Both segmentation networks use the bounding boxes of ground-truth masks as prompts. Areas marked in black in segmentation results represent non-planar regions.}
    \label{fig:SAMFailures}
\end{figure}

Some foundation models (or pretrained models) based on large-scale RGB image datasets, such as the Segment Anything Model (SAM)~\cite{SAM2023} and the Contrastive Language-Image Pretraining model (CLIP)~\cite{CLIP2021}, have shown excellent performance in the field of RGB image processing. 
If foundation models developed for three-band RGB image segmentation can be adapted to four-band RGB-D data, it is expected that a highly generalizable network for segmenting plane instances from RGB-D data can be trained cost-effectively. 
\textbf{However, it is not a matter of simply changing the input of a deep learning model from three bands to four bands.}
For example, if we were to fine-tune the original SAM into a model that can segment plane instances from RGB-D data, we would find that just fine-tuning SAM is also very expensive. 
Additionally, its inference speed is slow. 
If we fine-tune an accelerated SAM variant, like EfficientSAM~\cite{Xiong_CVPR2024}, we find that simply changing the model's input from three to four bands and then directly fine-tuning the original EfficientSAM's backbone results in limited generalization performance (\textbf{the first issue}; see Figure~\ref{fig:SAMFailures}). 
This is because the backbone of EfficientSAM, although simplified compared to the original SAM, remains complex. 
In addition, the properties of the D band differ significantly from those of the RGB bands. 
Therefore, to effectively learn D-band features using such a complex network structure, a large volume of training data is required. 
In other words, unlike the relatively straightforward transfer learning from an RGB image network to another RGB image network, transfer learning from an RGB image network to an RGB-D image network often requires significantly more training data for fine-tuning. 
However, all existing available training sets for RGB-D plane instance segmentation are limited in size (compared to the training sets of foundation models such as SAM and CLIP). 
Therefore, directly fine-tuning the backbone of the original EfficientSAM tends to result in overfitting. 
Additionally, EfficientSAM relies on input prompts (such as points or boxes) along with input images to predict masks, which prevents it from achieving fully automatic plane instance segmentation (\textbf{the second issue}).

The second issue is relatively easy to solve, as we can adopt an existing object detection network with superior performance, such as the classic Faster R-CNN~\cite{Ren_TPAMI2017}, as a plane detection network. 
We can then use the predicted bounding boxes as prompts for our modified EfficientSAM, thereby fully automating the plane instance segmentation task.

However, the first issue is relatively difficult to address. 
\textbf{Firstly}, we must solve the problem of how to introduce the D band into EfficientSAM. 
As mentioned above, keeping the backbone structure of EfficientSAM unchanged and only adding a D-band input would not yield an optimal solution. 
Abandoning EfficientSAM and instead directly using a much simpler model can effectively alleviate overfitting. 
However, this approach cannot fully utilize the feature representations learned by EfficientSAM from a large volume of RGB data. 
Therefore, we designed a dual-complexity backbone, where a simple convolutional neural network is primarily used to learn relevant feature representations of the D band, while the original complex backbone structure of EfficientSAM is mainly utilized for the RGB bands. 
This backbone structure helps avoid overfitting caused by the limited size of existing RGB-D plane segmentation datasets, while fully utilizing the powerful feature representation capabilities of EfficientSAM for the RGB bands, thereby comprehensively using all bands of the RGB-D data. 
Furthermore, the original backbone branch is not frozen so that it can be better adapted to the current task. 
Note that if the complexity of the dual-branch backbone is not carefully designed like our dual-complexity backbone, but we still want to retain the feature representations of its original branch, the original branch should be frozen, as in DPLNet~\cite{DPLNet2023}. 
Otherwise, the well-trained feature representations of the original branch would be damaged by subsequent fine-tuning, and with limited fine-tuning data, it would easily result in overfitting. 
\textbf{Secondly}, the domain of RGB-D bands significantly differs from that of RGB bands. 
Therefore, directly fine-tuning our modified EfficientSAM on existing limited-scale RGB-D plane instance segmentation datasets is likely to result in suboptimal performance. 
To ensure that our dual-complexity backbone also has a strong generalization ability when processing RGB-D data, we collected approximately 100,000 RGB-D images and used SAM-H~\cite{SAM2023} to perform fully automatic mask annotation using only the RGB bands for the segment anything task. 
We first pretrained our PlaneSAM on this large-scale, imperfect RGB-D dataset in a self-supervised manner, and then fine-tuned it specifically for the plane instance segmentation task. 
\textbf{Thirdly}, EfficientSAM does not perform well when segmenting large-area planes, as its loss function favors only the segmentation of small regions. 
To solve this problem, we optimized the combination ratio of loss functions so that the model can handle both large- and small-sized plane instance segmentation effectively.

To our knowledge, \textbf{we are the first to utilize all four bands of RGB-D data in a deep learning model to segment plane instances}. The following are the specific contributions of this study:

\begin{itemize}[leftmargin=*, label=\scalebox{1.2}\textbullet, itemsep=0em]  
    \item We enhanced plane instance segmentation with the depth band by designing a novel dual-complexity backbone structure. 
    Unlike existing dual-branch backbones, it can avoid overfitting caused by the limited scale of existing RGB-D plane instance segmentation datasets, fully utilize the feature representation capabilities of the foundation model EfficientSAM, and also allows the original backbone branch to be fine-tuned for better adaptation to the new task.

    \item A self-supervised pretraining strategy based on imperfect pseudo-labels is used to enable cost-effective pretraining of our PlaneSAM on large-scale RGB-D data.
    Different from traditional costly pretraining that relies on fine-grained manual annotations~\cite{He_TPAMI2020, DINO2023}, we used the imperfect segmentation results of RGB-D data automatically produced by SAM-H to pretrain our plane segmentation model. 
    Our pretraining success indicates that the imperfect pseudo-labels produced by algorithms for related tasks involving the same type of data can still contribute to model training. 

    \item We optimized the loss function combination ratio of EfficientSAM so that the fine-tuned PlaneSAM can effectively segment large planes from  RGB-D data.

    \item Our PlaneSAM increases the computational overhead by only approximately 10\% compared to EfficientSAM, while achieving state-of-the-art performance on the ScanNet~\cite{Liu_CVPR2018}, Matterport3D~\cite{Chang_3DV2017}, ICL-NUIM RGB-D~\cite{Handa_ICRA2014}, and 2D-3D-S~\cite{2D3DS_2017} datasets.
\end{itemize}

\section{Related works}
\label{sec:Related_Works}

\subsection{Plane segmentation from RGB-D data}  
\label{Sec:PlaneSeg}
As deep learning has advanced rapidly, methods based on it have gradually dominated the field of plane instance segmentation from a single-view RGB-D image due to their high accuracy and robustness. 
PlaneNet~\cite{Liu_CVPR2018} adopts a top-down approach, directly predicting plane instances and 3D plane parameters from a single RGB image, achieving good results. 
However, this method has some drawbacks, such as the need to know the maximum number of planes in the image in advance and difficulties in handling small-area planes. 
In subsequent work, PlaneRCNN~\cite{Liu_CVPR2019} and PlaneAE~\cite{Yu_CVPR2019} address these issues. 
PlaneRCNN also adopts a top-down approach, using Mask R-CNN~\cite{He_TPAMI2020} as the plane detection network, allowing it to detect any number of planes. 
Then, PlaneRCNN uses a refinement network to jointly refine all predicted plane instance masks, ultimately obtaining globally refined segmentation masks. 
In contrast to PlaneRCNN, PlaneAE adopts a bottom-up approach. 
It first uses a CNN to learn embeddings for each pixel, and then groups the embeddings using an efficient mean shift clustering algorithm to obtain plane instances. 
PlaneTR~\cite{Tan_ICCV2021} explored how geometric cues affect plane instance segmentation and found that line segments contain more comprehensive 3D information. 
It uses a Transformer~\cite{Transformer2017} model to encode the contextual features of the input image and line segments into two sets of tokenized sequences. 
These sequences, along with a set of learnable plane queries, are input into a plane decoder to output a set of plane instances. 
PlaneAC~\cite{Zhang_PR2024} also uses a line-guided approach but adopts a hybrid model that combines Transformer and CNN. 
BT3DPR~\cite{Ren_GM2024} uses a bilateral Transformer to enhance the segmentation of small-sized planes. 
PlaneSeg~\cite{Zhang_TNNLS2024} developed CNN-based plugins to address the challenges of small-sized plane segmentation and improve boundary precision during plane segmentation. 
However, despite using RGB-D datasets for experiments, these methods rely solely on single RGB image inputs. Further research is needed to effectively utilize all four bands of RGB-D data for high-quality plane segmentation, which is the focus of this paper.

\subsection{Segment Anything Models}  
\label{Sec:SAM}
In the past few years, foundation models have made groundbreaking advancements. 
Models such as CLIP~\cite{CLIP2021}, GPT-4~\cite{GPT4_2024}, and BERT~\cite{Devlin_2019Bert} have made significant contributions across various fields. 
In computer vision, the Segment Anything Model (SAM)~\cite{SAM2023} is a highly significant large interactive vision model, which has been proven to be able to segment any object given some appropriate prompts in advance. 
It consists of three modules: 1) an image encoder, which directly adopts the design of ViT~\cite{ViT2021} to process the image into intermediate image features; 2) a prompt encoder, which supports points, boxes and masks inputs, and converts the input prompts into embedding tokens; and 3) a mask decoder, which uses the cross-attention mechanism to effectively interact with the image features and the prompt embedding tokens to generate the final mask. 
SAM has been widely used in tasks such as semantic segmentation and instance segmentation due to its powerful zero-shot generalization ability. 
However, it suffers from the major problem of high computational cost, which makes real-time processing challenging.

To solve the problem of low efficiency, researchers have developed several smaller and faster SAM models, such as Fast SAM~\cite{FastSAM2023}, Mobile SAM~\cite{MobileSAM2023} and EfficientSAM~\cite{Xiong_CVPR2024}. 
Among them, EfficientSAM achieves the best results, so we study its application in plane instance segmentation from RGB-D data. 
EfficientSAM adopts masked mask image pretraining to learn effective visual representations by reconstructing features from the SAM image encoder, and then fine-tunes with the SA-1B dataset established by Meta AI Research~\cite{SAM2023}, achieving notable results. 
With the smaller ViT backbone, EfficientSAM decreases the parameter count and enhances inference speed by approximately 20 times compared to SAM, while the accuracy is only reduced by about 2\%. 
However, there are still some problems that cannot be ignored when introducing EfficientSAM into the RGB-D data plane instance segmentation task. 
First, EfficientSAM is still a very complex network, and the domain of RGB images differs significantly from that of RGB-D images, which means we need a large number of RGB-D plane segmentation samples to fine-tune the network. 
However, the current RGB-D plane segmentation dataset is relatively small. 
Therefore, we must design appropriate network architectures and training strategies; otherwise, it is easy for the network, at least the part concerning the D band, to fall into overfitting. Secondly, EfficientSAM's performance in segmenting large-area planes is unsatisfactory. 
Before fine-tuning, EfficientSAM is greatly affected by changes in illumination and color, and it tends to divide parts of the same plane with different illumination or colors into different plane instances. 
After fine-tuning using RGB-D plane segmentation datasets, however, EfficientSAM still cannot effectively segment large planes because its loss function tends to favor segmenting small objects. 
Thirdly, EfficientSAM predicts masks using input prompts, like points and boxes, in addition to input images, which makes it unable to achieve fully automatic plane instance segmentation.
This paper will provide reasonable solutions to all of the aforementioned issues.

\section{Method}
\label{sec:Method}
\begin{figure}[h]
   \centering
   {
   \includegraphics[width=146.5mm,height=75mm]{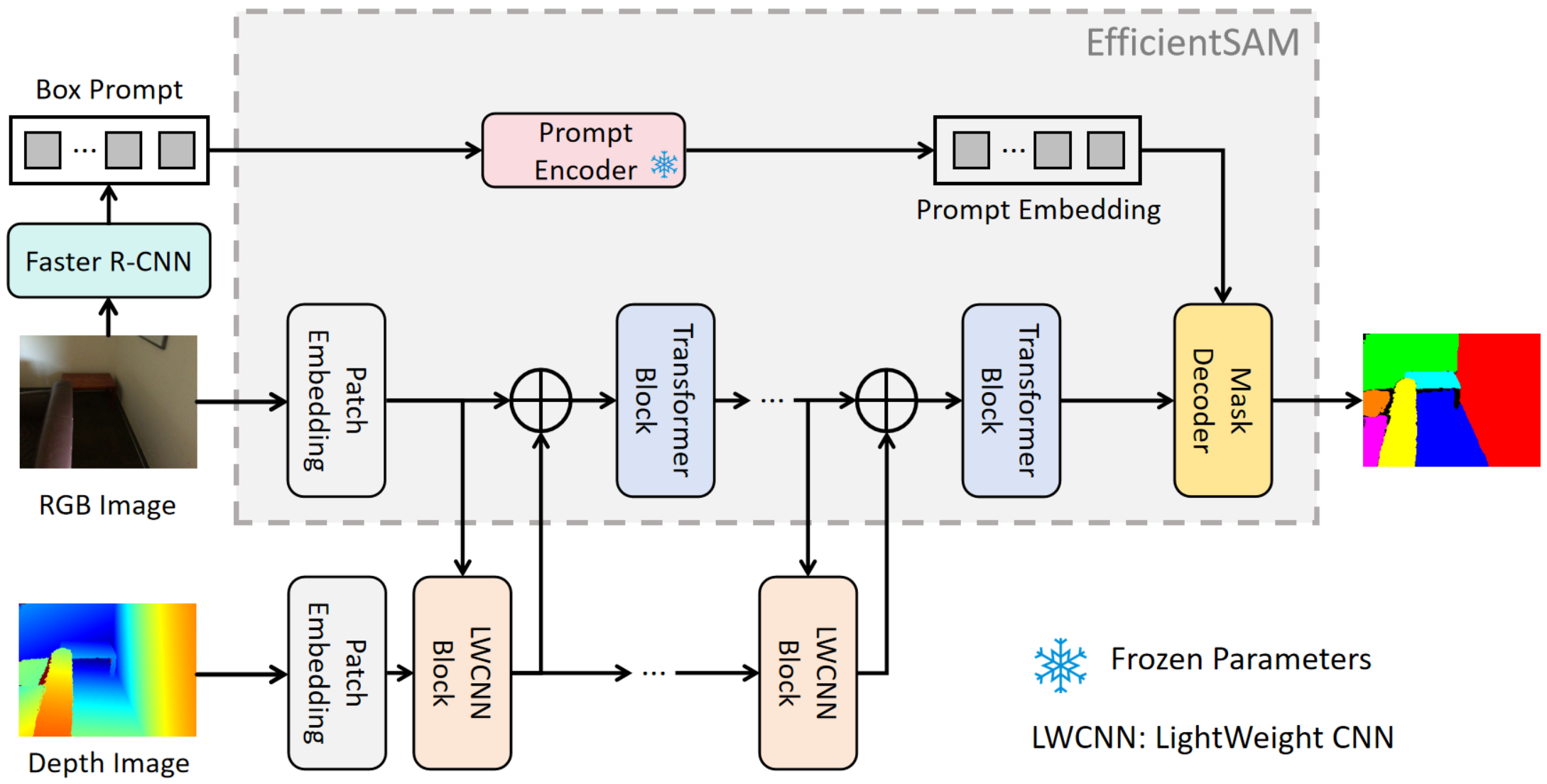} }
    \captionsetup{font=small}  
    \caption{Detailed schematic of our PlaneSAM.}
   \label{fig:PlanSAM}

\end{figure}

We use a top-down approach to segment plane instances from RGB-D data in a multimodal manner. Initially, we detect planes in the data and obtain their bounding boxes, which are then input into the modified multimodal EfficientSAM~\cite{Xiong_CVPR2024} as prompts to generate masks for the corresponding plane instances. Figure~\ref{fig:PlanSAM} illustrates that our network comprises two subnetworks: a plane detection network and a multimodal plane instance mask generation network. Because our method is primarily based on EfficientSAM for plane instance segmentation, we refer to it as PlaneSAM. The plane detection network, the multimodal plane instance mask generation network, and the adopted loss function are introduced in Sections 2.1, 2.2, and 2.3, respectively. Section 2.4 provides a detailed description of the methods used to pretrain our PlaneSAM on the segment anything task using the RGB-D datasets and fine-tune it on the ScanNet dataset~\cite{Liu_CVPR2018}.

\subsection{Plane detection network}  
\label{Sec:Plane_detection}
Because the present study does not focus on plane detection, we do not attempt to make improvements in this regard. 
Similar to PlaneRCNN~\cite{Liu_CVPR2019}, we use the classic Faster R-CNN~\cite{Ren_TPAMI2017} for plane detection. 
We consider plane instances as targets, with Faster R-CNN needing only to predict two categories: plane and non-plane. 
We do not modify the components of Faster R-CNN; the only change we make is to enable it to predict the bounding boxes of plane instances. 
In our experiments, we found that simply using all the pretrained weights would significantly reduce the accuracy of Faster R-CNN; accordingly, we train all components of the network from scratch other than the pretrained ResNet backbone.

Please note that other object detection methods also can be modified for plane detection as we did with Faster R-CNN, and we could also potentially design a specialized plane detector for RGB-D data. Both approaches could possibly yield better results than Faster R-CNN. However, we adopt this classical object detector to demonstrate that the superior segmentation results achieved by our PlaneSAM do not rely on perfect plane detection outcomes.

\subsection{Plane instance mask generation network}  
\label{Sec:Plane_mask}
To enhance plane instance segmentation with the depth band (the geometric band), we design a dual-complexity backbone structure to learn and fuse the feature representations of RGB and D bands. 
Because a large volume of annotated RGB image segmentation data is available, we primarily use the complex backbone of EfficientSAM to learn the feature representations of RGB bands. 
EfficientSAM~\cite{Xiong_CVPR2024} uses the Vision Transformer (ViT)~\cite{ViT2021} as its image encoder, which initially converts an input image into patch embeddings, then extracts image features through multiple encoder blocks (Transformer blocks), and finally outputs the feature embeddings. Following the traditional Transformer structure~\cite{Transformer2017}, each encoder block comprises a multi-head attention layer and a feedforward neural network. 
Because it is a large Transformer model trained using a substantial set of RGB images, EfficientSAM achieves strong generalization capabilities.

Theoretically, directly adding a D-band input and performing fine-tuning may enable EfficientSAM to segment plane instances from RGB-D data. 
However, as analyzed in the Introduction, this approach is highly vulnerable to overfitting, particularly in the D-band portion. 
This underutilizes the powerful capabilities of EfficientSAM. Abandoning the use of complex EfficientSAM and directly switching to a simpler network model can effectively alleviate overfitting. 
However, the limited capabilities of such a network would result in suboptimal performance. 
To solve this problem, we adopt a dual-complexity backbone structure. Specifically, the backbone primarily uses a simple CNN branch to learn feature representations of the D band and a complex Transformer branch (the backbone of EfficientSAM) to learn feature representations of the RGB bands. 
This approach helps avoid the overfitting problem stemming from the limited availability of RGB-D plane instance segmentation data, fully utilizes the powerful feature representation capabilities of the foundational model EfficientSAM, while allowing the original complex Transformer branch to be moderately fine-tuned, thereby achieving the comprehensive information fusion of all four bands of RGB-D data.

For the simple CNN branch, we employ the multimodal prompt generator module from DPLNet~\cite{DPLNet2023}, which is originally designed for semantic segmentation, as the basic CNN block. 
Of course, other lightweight CNN (LWCNN) block designs can also be adopted because the key point in selecting these CNN blocks is that each of them should be significantly simpler than the Transformer block in the backbone of EfficientSAM. 
Therefore, the structures of these CNN blocks can even be different from one another. 
We choose such a CNN block design because in addition to being a LWCNN, it can fuse RGB- and D-band features at multiple levels, enabling more robust feature representation learning. 
We add a LWCNN block before each encoder block of EfficientSAM. 
The RGB and D bands are first input into a group of EfficientSAM's encoder block and LWCNN block. 
Except for the first group, each group receives the output of the previous group and further fuses features. 
Different from DPLNet~\cite{DPLNet2023}, we use only the output of the last group as input for the mask decoder, instead of inputting the output of each group into the mask decoder. 
Furthermore, we do not freeze the original branch (the complex RGB Transformer branch in our paper) for the following two reasons:

1) We are transferring a network originally designed for the segment anything task to the plane instance segmentation task, and the two tasks differ significantly. Therefore, we need to fine-tune the original RGB network branch (the backbone of EfficientSAM) to fully adapt it to plane instance segmentation.

2) Not freezing the original backbone branch will not result in overfitting in our network setting, because it is still mainly the simple CNN branch that is learning the features of the D band. It can be explained as follows. In this paper, the complex network branch adopts a Transformer architecture. As shown in Figure~\ref{fig:PlanSAM}, 11 Transformer blocks are connected in parallel to the CNN branch consisting of 12 CNN blocks. Because the complexity of the 11 Transformer blocks is much higher than that of the 12 CNN blocks, the learning speed of the former is significantly lower than that of the latter under an equivalent learning rate. Therefore, when the CNN branch has already learned the main features of the D band, the weights of the 11 Transformer blocks do not change significantly. Thus, when the weights of the complex network branches are not frozen, our PlaneSAM still can avoid overfitting.

Therefore, although we use a dual-branch network structure like DPLNet~\cite{DPLNet2023}, our backbone structure is fundamentally different from theirs: they try to retain the pretrained features by forcefully freezing the weights of the original branch, whereas we use a dual-complexity backbone structure that not only preserves the feature representations of the original RGB bands but also allows fine-tuning of the original backbone branch to better adapt to new tasks, while primarily letting the newly added branch learn features of the new band. 
Experimental results also demonstrate that our dual-complexity backbone structure produces better plane segmentation results than the dual-branch network structure that simply freezes the original branch.

Beyond the application discussed in this paper, our dual-complexity backbone structure shows great potential for other tasks where we need to 
transfer a foundation model to a new domain, such as transferring an RGB foundation model to the RGB+X domain (where X may be the D band, the near-infrared band, etc.), with a relatively small fine-tuning dataset. 
At present, an increasing number of foundation models, such as SAM~\cite{SAM2023}, CLIP~\cite{CLIP2021}, and ChatGPT~\cite{GPT4_2024}, have been developed, and more are currently in development. 
We often need to extend these pretrained large models to other domains, but the fine-tuning data available is often limited in scale. 
Therefore, our dual-complexity backbone structure is expected to find more and more applications.

However, it should be noted that if the complexity of the two branches in other dual-branch backbones has not been carefully designed like our dual-complexity backbone, but the goal is still to retain the feature representations of the original branch, we recommend freezing the original branch, as in DPLNet~\cite{DPLNet2023}. 
Otherwise, the well-trained feature representations of the original branch could be damaged by subsequent fine-tuning, and if the fine-tuning training set is small, it could easily result in overfitting.

\subsection{Loss function}  
\label{Sec:Loss_func}
When used for plane instance segmentation, EfficientSAM~\cite{Xiong_CVPR2024} exhibits poor performance on large-area instances, even when it has been fine-tuned. 
This is mainly because EfficientSAM was originally designed to segment various types of objects, and small objects typically account for the majority of objects in an image containing multiple objects; consequently, EfficientSAM places higher focus on segmenting small objects, making it less effective when handling large areas.
In this paper, we address this issue by modifying the loss function.

To better elaborate on our setup of the loss function, we first review the standard loss function used to train the EfficientSAM. 
Focal loss~\cite{Focal2017} is designed to tackle class imbalance by reducing the weights of easily classified samples, enabling the model to allocate more attention towards difficult-to-classify samples. 
Consequently, focal loss performs well in the segmentation of small objects. 
However, because of the overemphasis on difficult-to-classify samples, which are often small objects or objects with unclear boundaries, the segmentation effect may be less effective for large-area instances. 
Dice loss~\cite{Dice2020} performs well in handling class imbalance and large-area instances by directly optimizing the evaluation metric (Dice coefficient) in the segmentation task. 
Dice loss treats minority and majority classes equally during calculation, making it suitable for problems involving unbalanced classes. 
Furthermore, Dice loss is more effective when handling large-area instances, as it considers the proportion of overlapping parts in its calculations. However, in some cases, the optimization process of Dice loss might not be stable, and numerical instability may occur owing to its denominator.

EfficientSAM uses a linear combination of focal, dice, and MSE losses with a 20:1:1 ratio. 
However, based on the above analysis of focal loss and Dice loss, we consider that this combination is not optimal for plane instance segmentation, as the weight of the focal loss is too large, causing the model to perform well when segmenting small objects, but poorly when processing large objects. 
In other words, the model focuses too much on small objects that are difficult to segment at the cost of a lower segmentation quality of large-area instances. 
In addition, the MSE loss is mainly used to train the IoU prediction head. 
Because we found that there is no need to specifically fine-tune the IoU prediction head for the plane instance segmentation task, we omitted the MSE loss.

To improve the effect of plane instance segmentation, we adjust the loss function to a 1:1 linear combination of the focal and Dice losses. 
By balancing these losses, the model will not exhibit bias towards segmenting small objects and can effectively handle large-area instances.
Our experimental results also show that combining the focal and Dice losses in a 1:1 ratio yields optimal performance.
Therefore, this combination maintains the effective segmentation of small objects while significantly improving the segmentation quality of large-area instances.

\subsection{Network training}  
\label{Sec:Network_training}
A two-step process is used to train our network. First, the dual-complexity backbone of our PlaneSAM is pretrained on RGB-D datasets for the segment anything task using the imperfect pseudo-labels generated by SAM-H~\cite{SAM2023}, and then it is fine-tuned on an RGB-D dataset for the plane instance segmentation task. 
Different from traditional pretraining that relies on manual fine annotations~\cite{He_TPAMI2020, DINO2023, DETR2020}, which is highly costly, we used the imperfect segmentation results of RGB-D data automatically produced by an existing model designed for a much different task to pretrain our plane segmentation network. 
Therefore, our pretraining is much more cost-effective.

\textbf{Pretraining.} The original EfficientSAM was trained on RGB images, whose domain is significantly different from that of RGB-D images. Therefore, directly fine-tuning our PlaneSAM on the existing RGB-D plane instance segmentation dataset---which contains a limited number of samples---would not sufficiently familiarize the model with the domain of RGB-D data, resulting in constrained training effectiveness. 
Although annotating a large-scale RGB-D plane instance segmentation training dataset may solve this issue, it would incur a very high cost.

To solve this problem, we pretrain our dual-complexity backbone for the segment anything task using RGB-D data.
In order to achieve good pretraining results, we need a large RGB-D training dataset for the segment anything task. 
However, there is no such a large, publicly available, manually annotated dataset. 
To address this issue, we used pseudo-labels automatically generated by SAM-H.
We collected approximately 100,000 RGB-D images from three datasets: ScanNet\_25k~\cite{Dai_CVPR2017}, SUN RGB-D~\cite{Song_CVPR2015}, and 2D-3D-S~\cite{2D3DS_2017}. 
Since SAM~\cite{SAM2023} was developed for RGB images, we feed only the RGB bands of RGB-D data into SAM-H to automatically generate masks, which are used as ground-truth to pretrain our PlaneSAM. 
Thus, this strategy enables self-supervised pretraining of our PlaneSAM on large-scale RGB-D data, adapting it to the domain of RGB-D data at a low cost. 
Although the quality of segmentation masks (pseudo-labels) automatically generated by SAM-H may not be very high, our experimental results show that our pretraining approach is highly effective.

\begin{figure}[htbp!]
    \centering
    \begin{minipage}{0.23\linewidth}
        \centering
        \includegraphics[width=\linewidth]{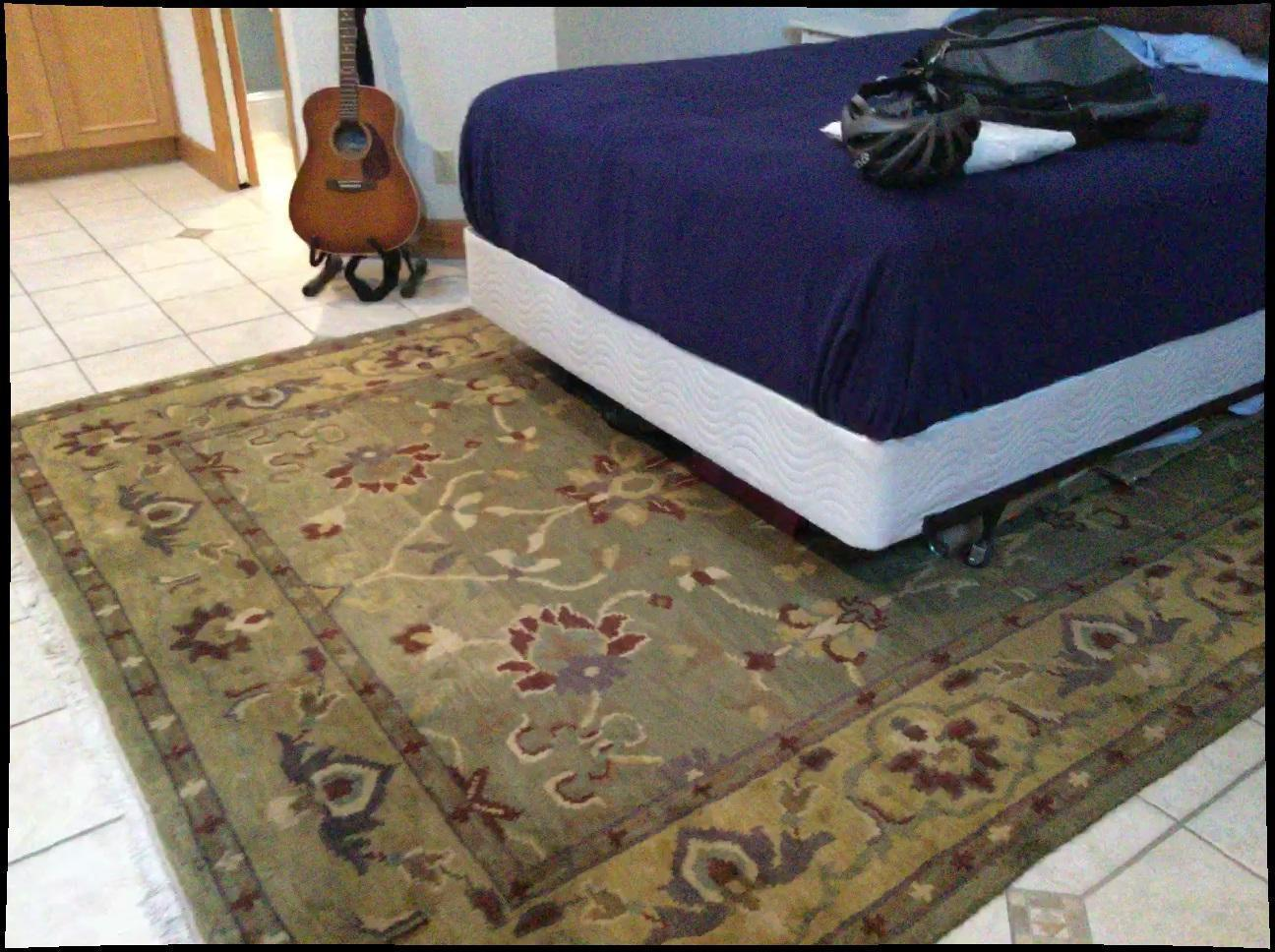}
    \end{minipage}%
    \hspace{0.05em} 
    \begin{minipage}{0.23\linewidth}
        \centering
        \includegraphics[width=\linewidth]{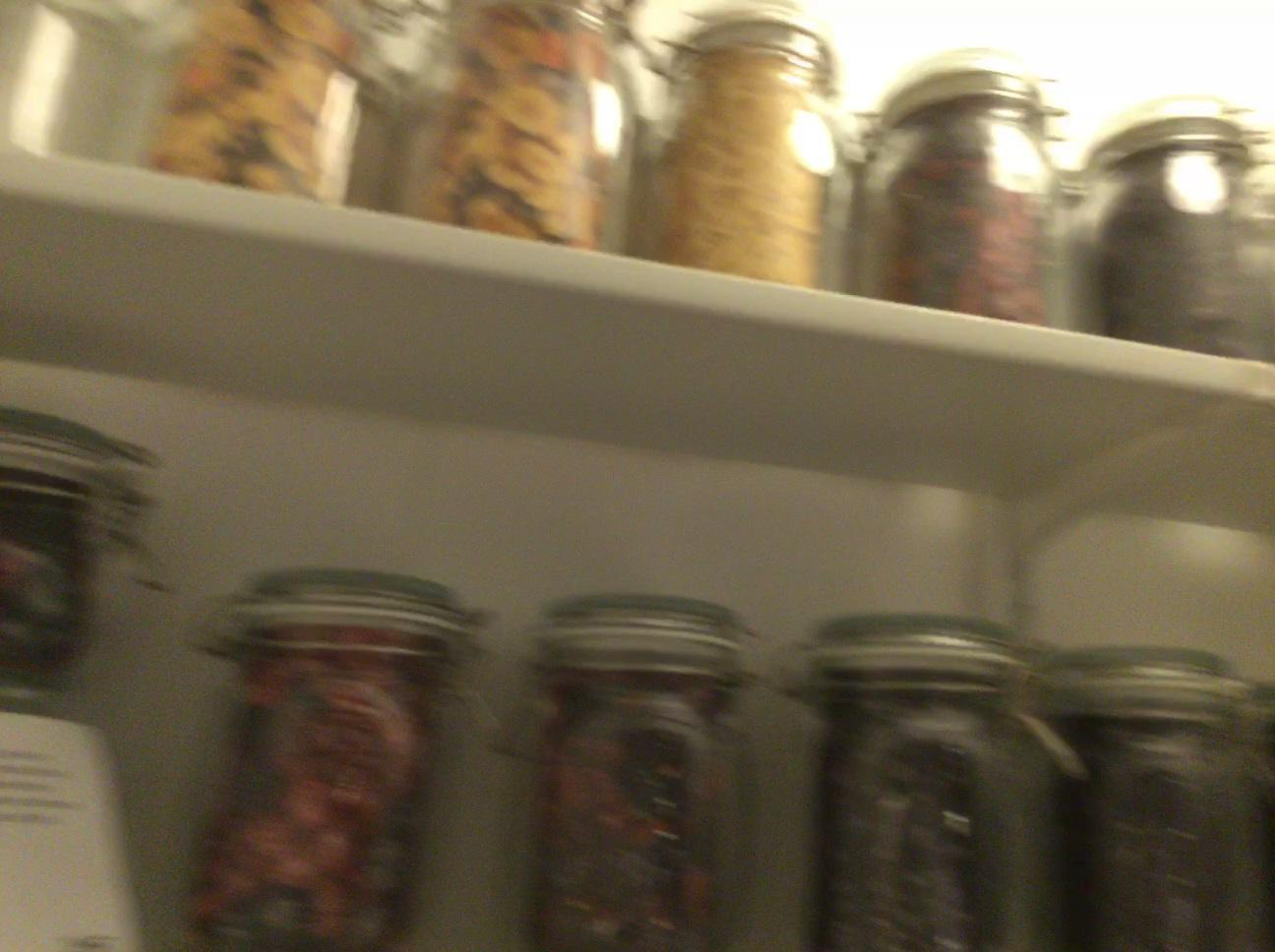}
    \end{minipage}%
    \hspace{0.05em} 
    \begin{minipage}{0.23\linewidth}
        \centering
        \includegraphics[width=\linewidth]{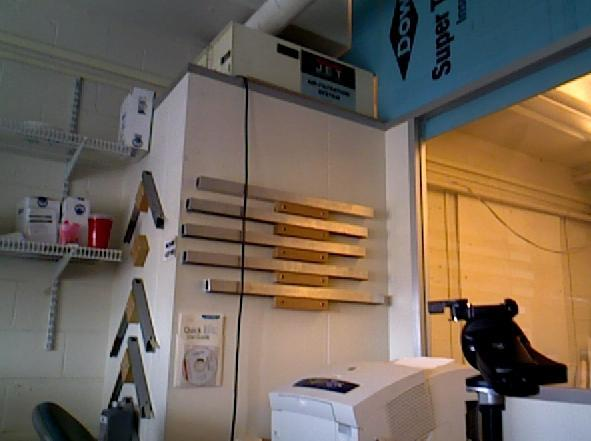}
    \end{minipage}%
    \hspace{0.05em}
    \begin{minipage}{0.23\linewidth}
        \centering
        \includegraphics[width=\linewidth]{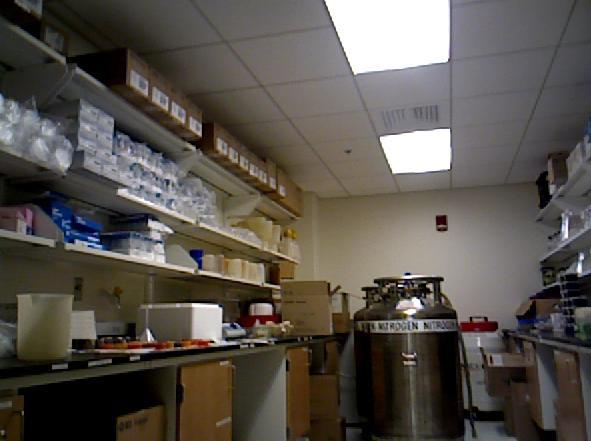}
    \end{minipage}

    \vspace{0.4em} 

    \begin{minipage}{0.23\linewidth}
        \centering
        \includegraphics[width=\linewidth]{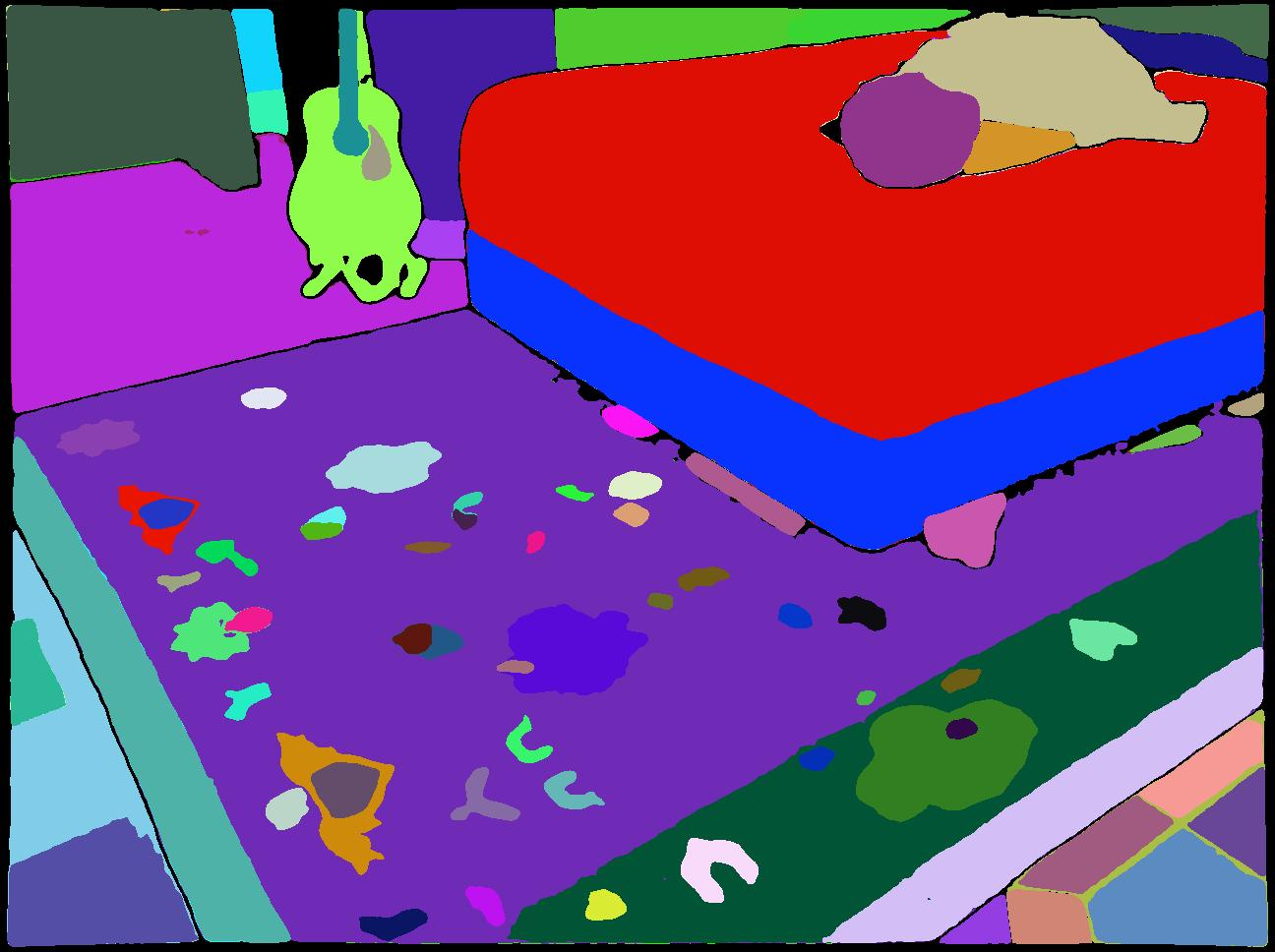}
    \end{minipage}%
    \hspace{0.05em} 
    \begin{minipage}{0.23\linewidth}
        \centering
        \includegraphics[width=\linewidth]{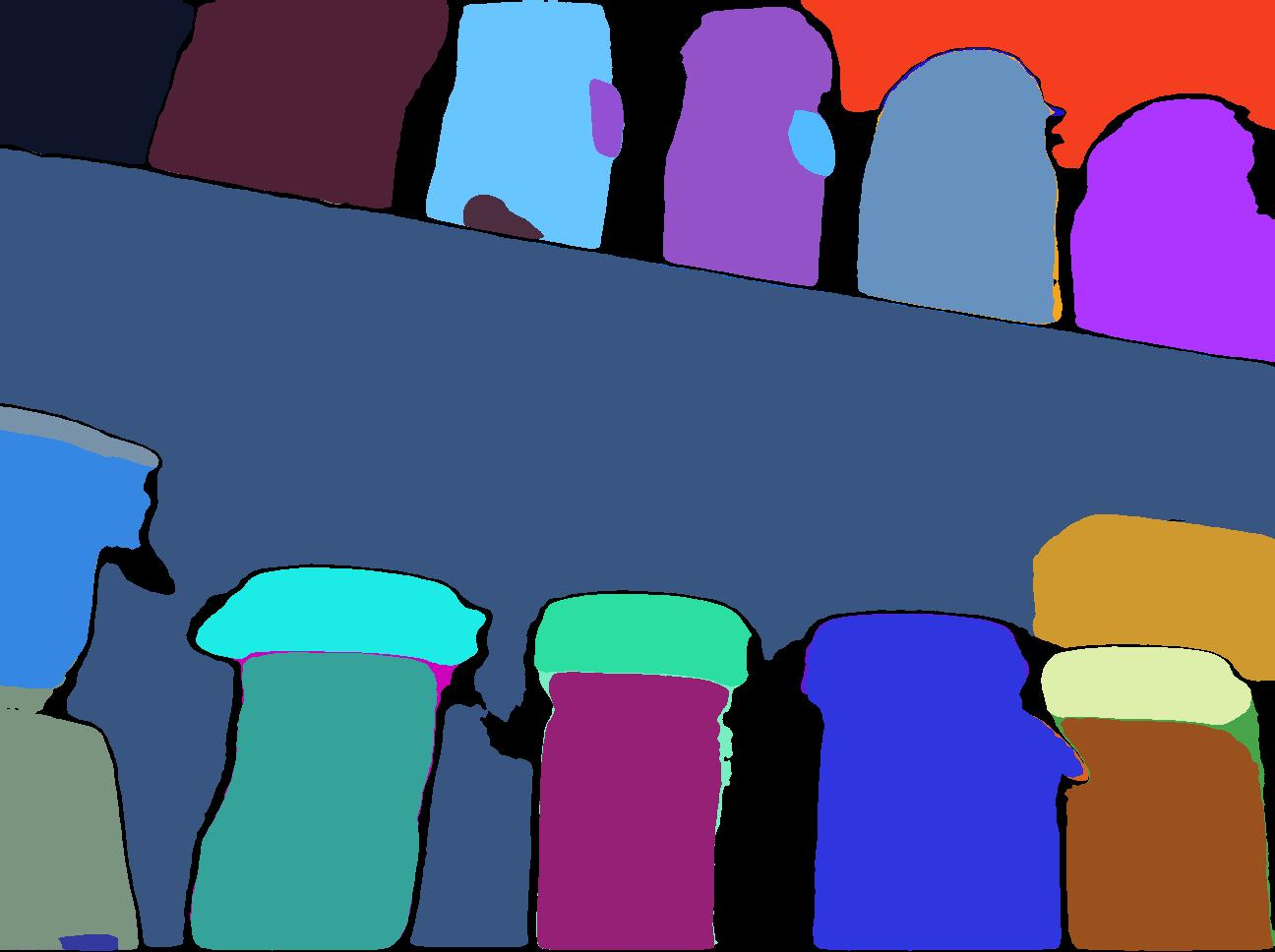}
    \end{minipage}%
    \hspace{0.05em} 
    \begin{minipage}{0.23\linewidth}
        \centering
        \includegraphics[width=\linewidth]{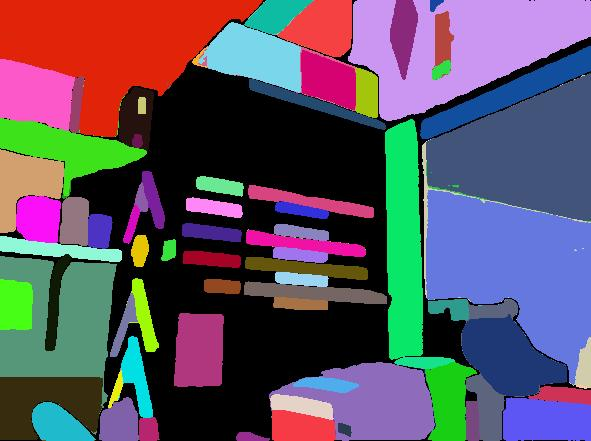}
    \end{minipage}%
    \hspace{0.05em} 
    \begin{minipage}{0.23\linewidth}
        \centering
        \includegraphics[width=\linewidth]{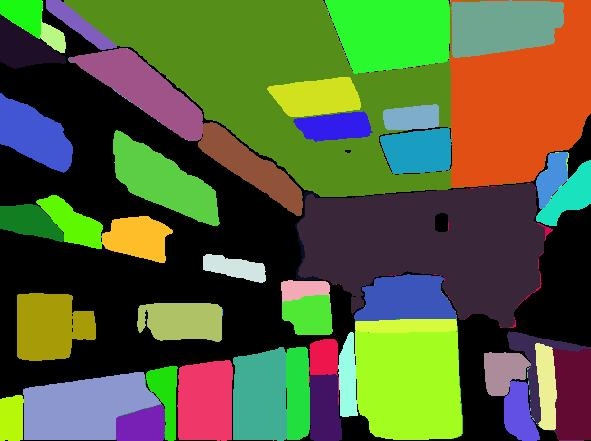}
    \end{minipage}

    \vspace{0.4em} 
    \captionsetup{font=small}  
    \caption{Examples of the datasets used to pretrain our PlaneSAM. The first row displays RGB-D data, while the second row displays corresponding pseudo-labels automatically generated by SAM-H.}
    \label{fig:Pretraining_dataset}
\end{figure}

We pretrain our dual-complexity backbone on 80,000 images and use the remaining 20,000 for validation. 
Figure~\ref{fig:Pretraining_dataset} shows some of the SAM-H's automatic mask annotations, where we can see that even for the segment anything task, the segmentation results are not perfect, and that the segment anything task differs greatly from our plane instance segmentation task. 
It should be noted that the masks here are automatically generated because we used SAM's segment everything mode. 
In other modes, SAM requires necessary prompts for segmentation, such as point prompts, box prompts, or text prompts, and therefore cannot automatically complete the segmentation task.

Because the prompt encoder has already been well trained, it is frozen, and only the image encoder and mask decoder are trained. During pretraining, we randomly select a mask from each RGB-D image as the foreground and use the bounding box of the ground-truth mask as a prompt. Notably, for each RGB image, SAM can predict approximately 100 masks. However, a large portion of these masks have very small areas, which essentially do not contribute to the training of our PlaneSAM. Therefore, such masks are filtered out. Our PlaneSAM achieves improved accuracy through pretraining, which suggests that imperfect results from algorithms for related tasks using the same type of data can still contribute to model training.

\textbf{Fine-tuning.} 
In addition to differences between segmentation tasks, the fine-tuning strategy slightly differs from the pretraining strategy. 
Firstly, to segment plane instances of arbitrary scales, we do not filter out small-area plane instances. Secondly, we add random noise of 0-10\% to the lengths of the ground-truth bounding boxes, to enhance our PlaneSAM's adaptability to the bounding boxes generated by Faster R-CNN, which may not be very precise. 
In addition, random horizontal flipping is used for data augmentation. 
Our PlaneSAM outputs three masks based on the input RGB-D data and prompt, and we compute the losses for these three masks against the ground-truth mask through a 1:1 linear combination of focal and Dice losses. 
We then apply backpropagation using the minimum of these three losses.

\section{Experiments \& analysis}
\label{sec:Experiments}
This section presents experiments on four datasets to validate our PlaneSAM's effectiveness, as well as ablation studies conducted to evaluate each component's contribution to the network. 
It should be pointed out that, unless otherwise stated, the EfficientSAM component in our PlaneSAM refers to EfficientSAM-vitt~\cite{Xiong_CVPR2024}.

\subsection{Datasets and metrics}  
\label{Sec:Datasets_metrics}
\textbf{Datasets.} We utilized a total of four datasets. The ScanNet dataset, annotated by Liu et al.~\cite{Liu_CVPR2018}, was used for both training and testing, whereas the Matterport3D~\cite{Chang_3DV2017}, ICL-NUIM RGB-D~\cite{Handa_ICRA2014}, and 2D-3D-S~\cite{2D3DS_2017} datasets were used only for model testing. It should be noted that the 2D-3D-S dataset was used to pretrain but not fine-tune our PlaneSAM. Moreover, due to the significant differences between the segment anything task (pretraining task) and the plane instance segmentation task, we can consider that the 2D-3D-S dataset was not used to train our PlaneSAM.

\textbf{Evaluation Metrics.} 
As with the SOTA algorithms~\cite{Tan_ICCV2021,Ren_GM2024,Zhang_PR2024}, the following three evaluation metrics were used: Rand index (RI), variation of information (VOI), and segmentation coverage (SC). 
Among these three evaluation metrics, a smaller VOI value, as well as larger RI and SC values, indicate a higher segmentation quality.

\subsection{Implementation details}  
\label{Sec:Implementation}
Our PlaneSAM was implemented in PyTorch, and trained with the Adam optimizer and a cosine learning rate scheduler. We pretrained our PlaneSAM on RGB-D datasets for the segment anything task for 40 epochs using two NVIDIA RTX 3090 GPUs. The initial learning rate was 1e-4, decaying to 0 in the final epoch. The batch size was 12, and the weight decay 0.01. The configuration for fine-tuning on the ScanNet dataset was the same as used in pretraining, and we trained for only 15 epochs.

As for Faster R-CNN, it was trained on the ScanNet dataset for 10 epochs using two NVIDIA RTX 3090 GPUs, the SGD optimizer, and a cosine learning rate scheduler. The initial learning rate was 0.02, decaying to 0 in the final epoch. The batch size was eight, the weight decay 1e-4, and the momentum 0.9.

\subsection{Results on the ScanNet dataset}  
\label{Sec:Results_ScanNet}
\begin{figure}[t!]
    \centering

    \begin{minipage}{0.05\linewidth}
        \centering
        \adjustbox{valign=c}{\rotatebox{90}{\fontsize{10pt}{10pt}\selectfont{RGB}}}
    \end{minipage}%
    \hspace{-0.4em} 
    \begin{minipage}{0.9\linewidth} 
        \centering
        \begin{subfigure}{0.161\linewidth}
            \centering
            \includegraphics[width=\linewidth]{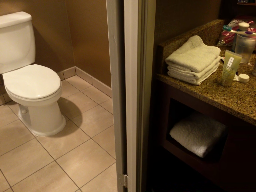}
        \end{subfigure}%
        \hfill 
        \begin{subfigure}{0.161\linewidth}
            \centering
            \includegraphics[width=\linewidth]{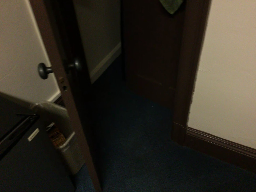}
        \end{subfigure}%
        \hfill
        \begin{subfigure}{0.161\linewidth}
            \centering
            \includegraphics[width=\linewidth]{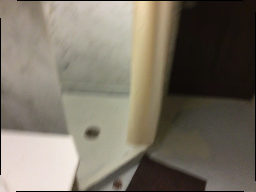}
        \end{subfigure}%
        \hfill
        \begin{subfigure}{0.161\linewidth}
            \centering
            \includegraphics[width=\linewidth]{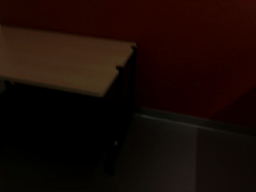}
        \end{subfigure}%
        \hfill
        \begin{subfigure}{0.161\linewidth}
            \centering
            \includegraphics[width=\linewidth]{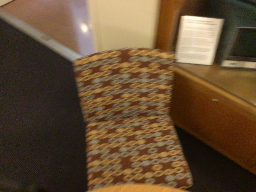}
        \end{subfigure}%
        \hfill
        \begin{subfigure}{0.161\linewidth}
            \centering
            \includegraphics[width=\linewidth]{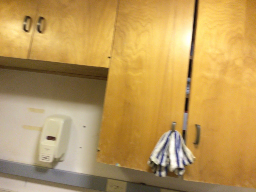}
        \end{subfigure}%
    \end{minipage}

    \vspace{0.15em} 
    \begin{minipage}{0.05\linewidth}
        \centering
        \adjustbox{valign=c}{\rotatebox{90}{\fontsize{10pt}{10pt}\selectfont{Depth}}}
    \end{minipage}%
    \hspace{-0.4em} 
    \begin{minipage}{0.9\linewidth} 
        \centering
        \begin{subfigure}{0.161\linewidth}
            \centering
            \includegraphics[width=\linewidth]{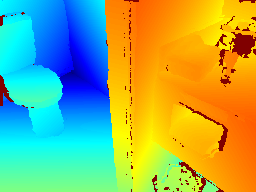}
        \end{subfigure}%
        \hfill 
        \begin{subfigure}{0.161\linewidth}
            \centering
            \includegraphics[width=\linewidth]{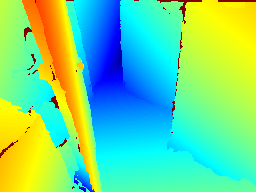}
        \end{subfigure}%
        \hfill
        \begin{subfigure}{0.161\linewidth}
            \centering
            \includegraphics[width=\linewidth]{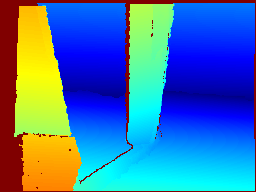}
        \end{subfigure}%
        \hfill
        \begin{subfigure}{0.161\linewidth}
            \centering
            \includegraphics[width=\linewidth]{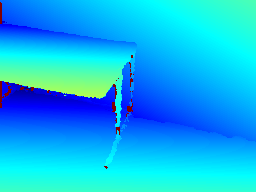}
        \end{subfigure}%
        \hfill
        \begin{subfigure}{0.161\linewidth}
            \centering
            \includegraphics[width=\linewidth]{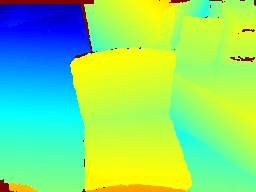}
        \end{subfigure}%
        \hfill
        \begin{subfigure}{0.161\linewidth}
            \centering
            \includegraphics[width=\linewidth]{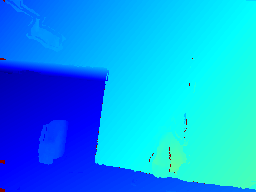}
        \end{subfigure}%
    \end{minipage}

\vspace{0.15em} 
    \begin{minipage}{0.05\linewidth}
        \centering
        \adjustbox{valign=c}{\rotatebox{90}{\fontsize{10pt}{10pt}\selectfont{PlaneAE}}}
    \end{minipage}%
    \hspace{-0.4em} 
    \begin{minipage}{0.9\linewidth} 
        \centering
        \begin{subfigure}{0.161\linewidth}
            \centering
            \includegraphics[width=\linewidth]{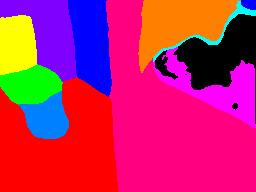}
        \end{subfigure}%
        \hfill 
        \begin{subfigure}{0.161\linewidth}
            \centering
            \includegraphics[width=\linewidth]{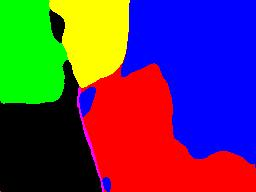}
        \end{subfigure}%
        \hfill
        \begin{subfigure}{0.161\linewidth}
            \centering
            \includegraphics[width=\linewidth]{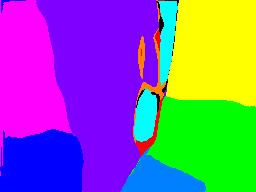}
        \end{subfigure}%
        \hfill
        \begin{subfigure}{0.161\linewidth}
            \centering
            \includegraphics[width=\linewidth]{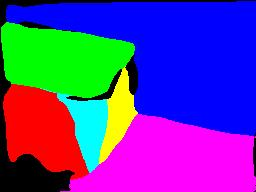}
        \end{subfigure}%
        \hfill
        \begin{subfigure}{0.161\linewidth}
            \centering
            \includegraphics[width=\linewidth]{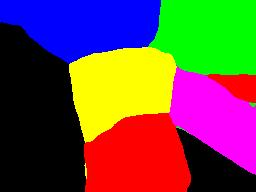}
        \end{subfigure}%
        \hfill
        \begin{subfigure}{0.161\linewidth}
            \centering
            \includegraphics[width=\linewidth]{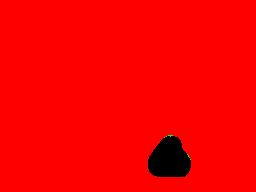}
        \end{subfigure}%
    \end{minipage}

\vspace{0.15em} 
    \begin{minipage}{0.05\linewidth}
        \centering
        \adjustbox{valign=c}{\rotatebox{90}{\fontsize{10pt}{10pt}\selectfont{PlaneTR}}}
    \end{minipage}%
    \hspace{-0.4em} 
    \begin{minipage}{0.9\linewidth} 
        \centering
        \begin{subfigure}{0.161\linewidth}
            \centering
            \includegraphics[width=\linewidth]{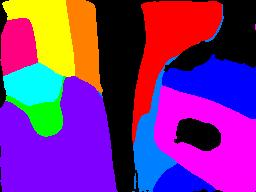}
        \end{subfigure}%
        \hfill 
        \begin{subfigure}{0.161\linewidth}
            \centering
            \includegraphics[width=\linewidth]{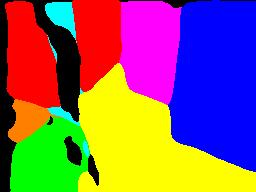}
        \end{subfigure}%
        \hfill
        \begin{subfigure}{0.161\linewidth}
            \centering
            \includegraphics[width=\linewidth]{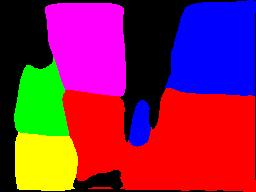}
        \end{subfigure}%
        \hfill
        \begin{subfigure}{0.161\linewidth}
            \centering
            \includegraphics[width=\linewidth]{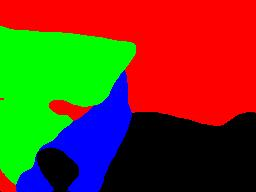}
        \end{subfigure}%
        \hfill
        \begin{subfigure}{0.161\linewidth}
            \centering
            \includegraphics[width=\linewidth]{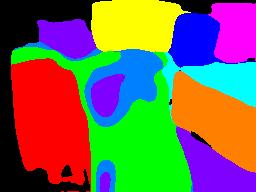}
        \end{subfigure}%
        \hfill
        \begin{subfigure}{0.161\linewidth}
            \centering
            \includegraphics[width=\linewidth]{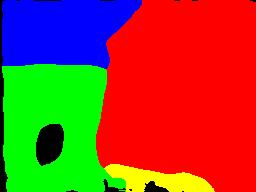}
        \end{subfigure}%
    \end{minipage}

\vspace{0.15em} 
    \begin{minipage}{0.05\linewidth}
        \centering
        \adjustbox{valign=c}{\rotatebox{90}{\fontsize{10pt}{10pt}\selectfont{X-PDNet}}}
    \end{minipage}%
    \hspace{-0.4em} 
    \begin{minipage}{0.9\linewidth} 
        \centering
        \begin{subfigure}{0.161\linewidth}
            \centering
            \includegraphics[width=\linewidth]{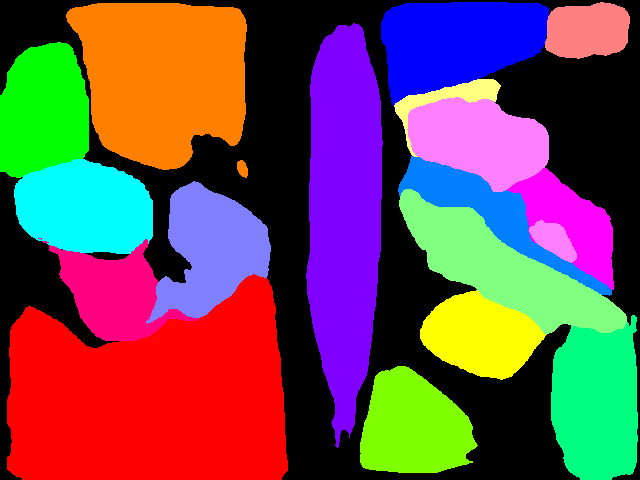}
        \end{subfigure}%
        \hfill 
        \begin{subfigure}{0.161\linewidth}
            \centering
            \includegraphics[width=\linewidth]{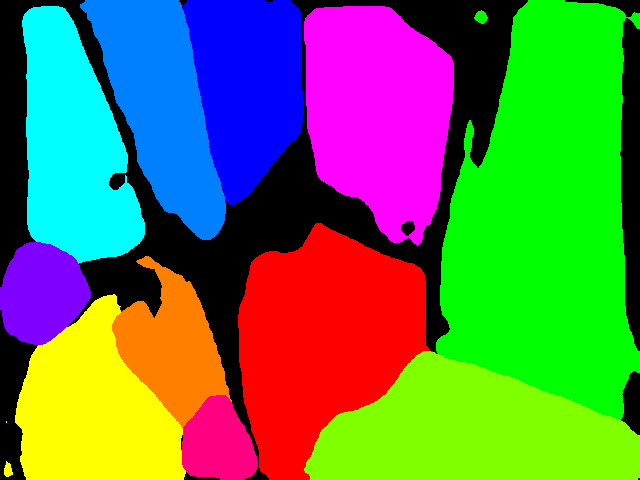}
        \end{subfigure}%
        \hfill
        \begin{subfigure}{0.161\linewidth}
            \centering
            \includegraphics[width=\linewidth]{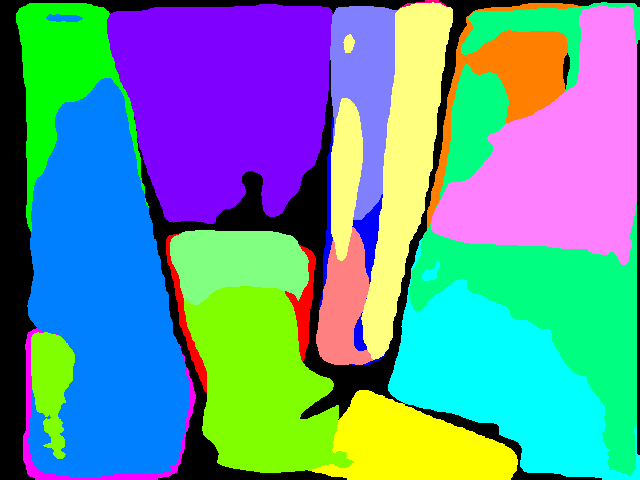}
        \end{subfigure}%
        \hfill
        \begin{subfigure}{0.161\linewidth}
            \centering
            \includegraphics[width=\linewidth]{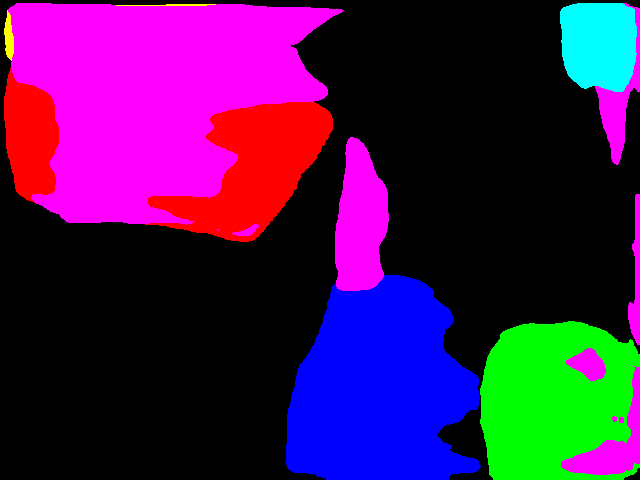}
        \end{subfigure}%
        \hfill
        \begin{subfigure}{0.161\linewidth}
            \centering
            \includegraphics[width=\linewidth]{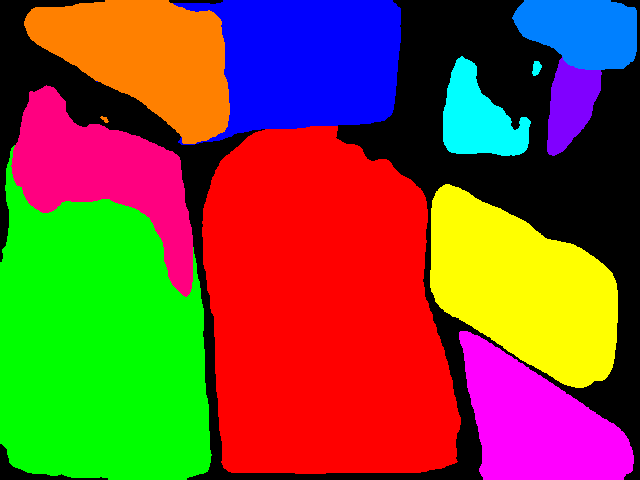}
        \end{subfigure}%
        \hfill
        \begin{subfigure}{0.161\linewidth}
            \centering
            \includegraphics[width=\linewidth]{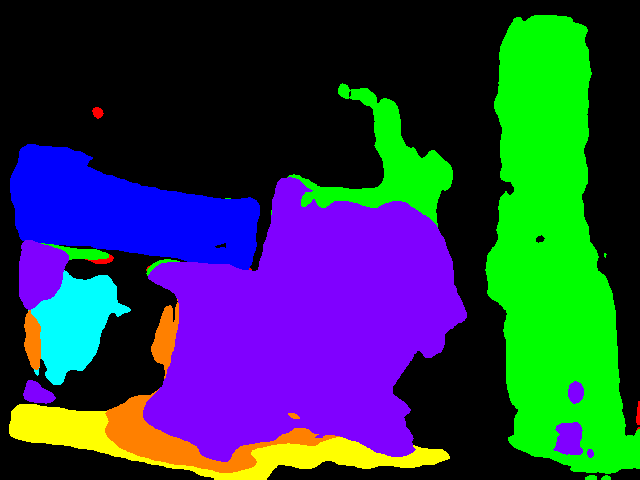}
        \end{subfigure}%
    \end{minipage}

\vspace{0.15em} 
    \begin{minipage}{0.05\linewidth}
        \centering
        \adjustbox{valign=c}{\rotatebox{90}{\fontsize{10pt}{10pt}\selectfont{Ours}}}
    \end{minipage}%
    \hspace{-0.4em} 
    \begin{minipage}{0.9\linewidth} 
        \centering
        \begin{subfigure}{0.161\linewidth}
            \centering
            \includegraphics[width=\linewidth]{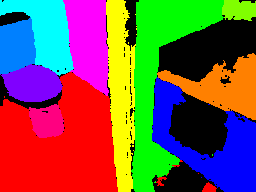}
        \end{subfigure}%
        \hfill 
        \begin{subfigure}{0.161\linewidth}
            \centering
            \includegraphics[width=\linewidth]{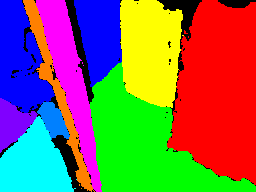}
        \end{subfigure}%
        \hfill
        \begin{subfigure}{0.161\linewidth}
            \centering
            \includegraphics[width=\linewidth]{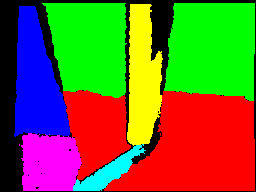}
        \end{subfigure}%
        \hfill
        \begin{subfigure}{0.161\linewidth}
            \centering
            \includegraphics[width=\linewidth]{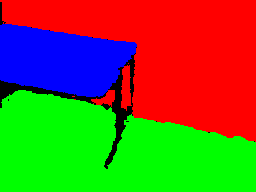}
        \end{subfigure}%
        \hfill
        \begin{subfigure}{0.161\linewidth}
            \centering
            \includegraphics[width=\linewidth]{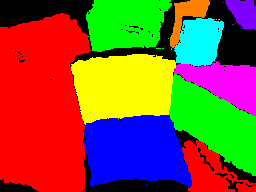}
        \end{subfigure}%
        \hfill
        \begin{subfigure}{0.161\linewidth}
            \centering
            \includegraphics[width=\linewidth]{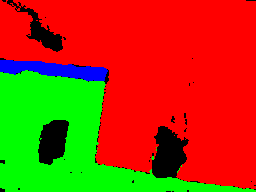}
        \end{subfigure}%
    \end{minipage}

\vspace{0.15em} 
    \begin{minipage}{0.05\linewidth}
        \centering
        \adjustbox{valign=c}{\rotatebox{90}{\fontsize{10pt}{10pt}\selectfont{GT}}}
    \end{minipage}%
    \hspace{-0.4em} 
    \begin{minipage}{0.9\linewidth} 
        \centering
        \begin{subfigure}{0.161\linewidth}
            \centering
            \includegraphics[width=\linewidth]{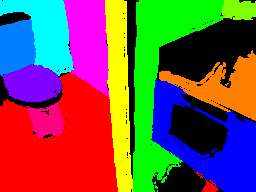}
        \end{subfigure}%
        \hfill 
        \begin{subfigure}{0.161\linewidth}
            \centering
            \includegraphics[width=\linewidth]{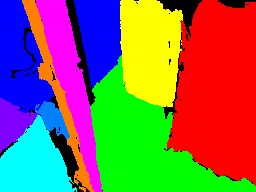}
        \end{subfigure}%
        \hfill
        \begin{subfigure}{0.161\linewidth}
            \centering
            \includegraphics[width=\linewidth]{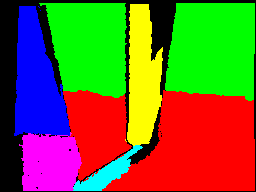}
        \end{subfigure}%
        \hfill
        \begin{subfigure}{0.161\linewidth}
            \centering
            \includegraphics[width=\linewidth]{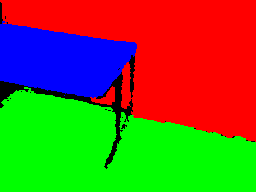}
        \end{subfigure}%
        \hfill
        \begin{subfigure}{0.161\linewidth}
            \centering
            \includegraphics[width=\linewidth]{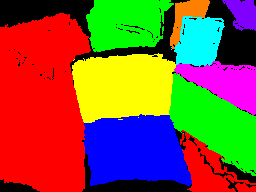}
        \end{subfigure}%
        \hfill
        \begin{subfigure}{0.161\linewidth}
            \centering
            \includegraphics[width=\linewidth]{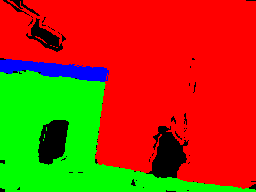}
        \end{subfigure}%
    \end{minipage}
    
    \vspace{0.4em} 
    \captionsetup{font=small}  
    \caption{Qualitative comparison results for some of the test algorithms on the ScanNet dataset. GT refers to ground-truth.}
    \label{fig:Results_Scannet}
\end{figure}

The proposed method was qualitatively and quantitatively compared with the following SOTA algorithms on the ScanNet dataset: PlaneAE~\cite{Yu_CVPR2019}, PlaneTR~\cite{Tan_ICCV2021}, X-PDNet~\cite{Cao_BMVC2023}, BT3DPR~\cite{Ren_GM2024}, and PlaneAC~\cite{Zhang_PR2024}. 
Our PlaneSAM used the same training and test sets as those of the baseline algorithms. 
X-PDNet was retrained from scratch by us to ensure that all tested methods use the same training set, whereas the other compared models were trained by their respective authors. 
Figure~\ref{fig:Results_Scannet} presents visual results of part of the test algorithms for plane instance segmentation on the ScanNet dataset, with our PlaneSAM notably exhibiting the best results. 
Table~\ref{tab:Results_ScanNet} presents the quantitative evaluation results of all test algorithms, further demonstrating that our method outperforms baseline methods across all metrics. 
Since none of the comparison algorithms use information from the D band, the comparative experiments in this section demonstrate that using all four bands of RGB-D data for plane segmentation is more effective than using only the RGB bands. 
Note that only some of the test algorithms' results are presented in Figure~\ref{fig:Results_Scannet}, as their official implementations are available to us, while others are not; the same applies to Figure~\ref{fig:Results_unseen}.  
Table~\ref{tab:Results_ScanNet} highlights the best results in bold, and the same applies to the following tables in this paper.

\begin{table}[htbp]
    \centering
    \small  
    \captionsetup{font=small}  
    \caption{Quantitative evaluation results on the ScanNet dataset.}  
    \label{tab:Results_ScanNet}
    \begin{tabular}{>{\centering\arraybackslash}m{3.6cm} >{\centering\arraybackslash}m{2.7cm} >{\centering\arraybackslash}m{2.7cm} >{\centering\arraybackslash}m{2.7cm}} 
        \toprule
        \multirow{2}{*}{{Method}} & \multicolumn{3}{c}{{ScanNet}} \\
        & {VOI} $\downarrow$ & {RI} $\uparrow$ & {SC} $\uparrow$ \\
        \midrule
        PlaneAE & 1.025 & 0.907 & 0.791 \\
        PlaneTR & 0.767 & 0.925 & 0.838 \\
        X-PDNet & 1.831 & 0.789 & 0.623 \\
        BT3DPR & 0.762 & 0.923 & 0.839 \\
        PlaneAC & 0.658 & 0.934 & 0.852 \\
        Ours  & \multicolumn{1}{c}{\bfseries 0.550} & \multicolumn{1}{c}{\bfseries 0.941} & \multicolumn{1}{c}{\bfseries 0.873} \\
        \bottomrule
    \end{tabular}
\end{table}

\subsection{Results on unseen datasets}  
\label{Sec:Results_unseen}
\begin{figure}[t!]
    \centering

    \begin{minipage}{0.05\linewidth}
        \centering
        \adjustbox{valign=c}{\rotatebox{90}{\fontsize{10pt}{10pt}\selectfont{RGB}}}
    \end{minipage}%
    \hspace{-0.4em} 
    \begin{minipage}{0.9\linewidth} 
        \centering
        \begin{subfigure}{0.161\linewidth}
            \centering
            \includegraphics[width=\linewidth]{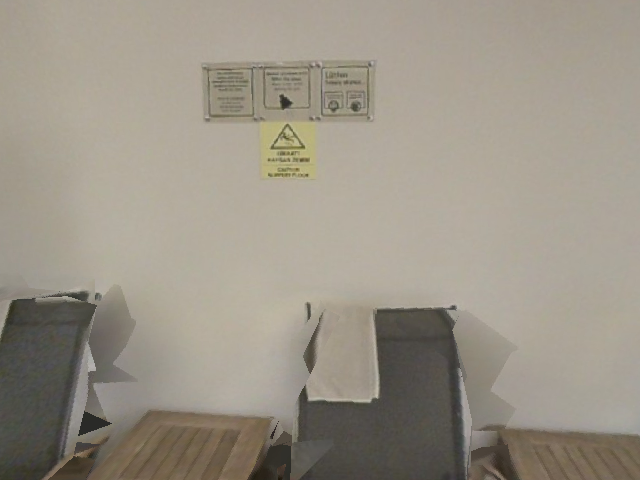}
        \end{subfigure}%
        \hfill 
        \begin{subfigure}{0.161\linewidth}
            \centering
            \includegraphics[width=\linewidth]{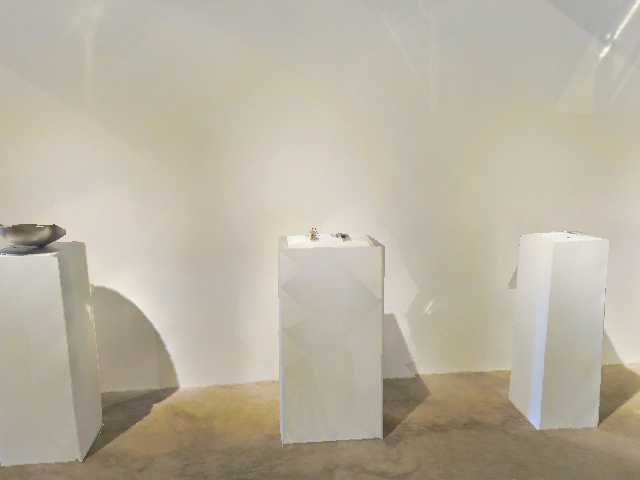}
        \end{subfigure}%
        \hfill
        \begin{subfigure}{0.161\linewidth}
            \centering
            \includegraphics[width=\linewidth]{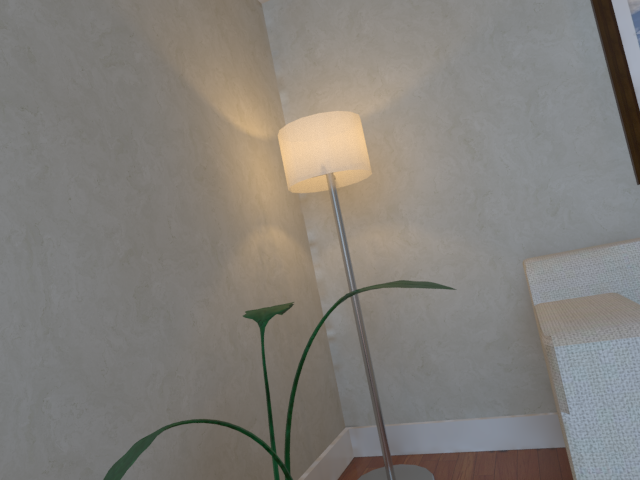}
        \end{subfigure}%
        \hfill
        \begin{subfigure}{0.161\linewidth}
            \centering
            \includegraphics[width=\linewidth]{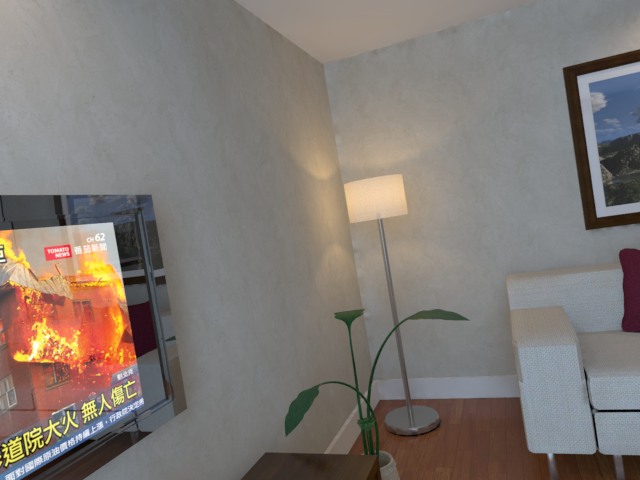}
        \end{subfigure}%
        \hfill
        \begin{subfigure}{0.161\linewidth}
            \centering
            \includegraphics[width=\linewidth]{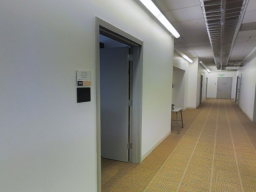}
        \end{subfigure}%
        \hfill
        \begin{subfigure}{0.161\linewidth}
            \centering
            \includegraphics[width=\linewidth]{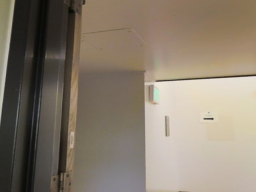}
        \end{subfigure}%
    \end{minipage}

    \vspace{0.15em} 
    \begin{minipage}{0.05\linewidth}
        \centering
        \adjustbox{valign=c}{\rotatebox{90}{\fontsize{10pt}{10pt}\selectfont{Depth}}}
    \end{minipage}%
    \hspace{-0.4em} 
    \begin{minipage}{0.9\linewidth} 
        \centering
        \begin{subfigure}{0.161\linewidth}
            \centering
            \includegraphics[width=\linewidth]{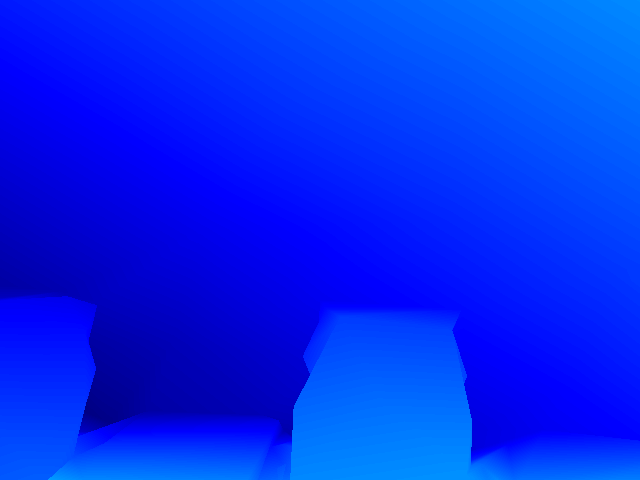}
        \end{subfigure}%
        \hfill 
        \begin{subfigure}{0.161\linewidth}
            \centering
            \includegraphics[width=\linewidth]{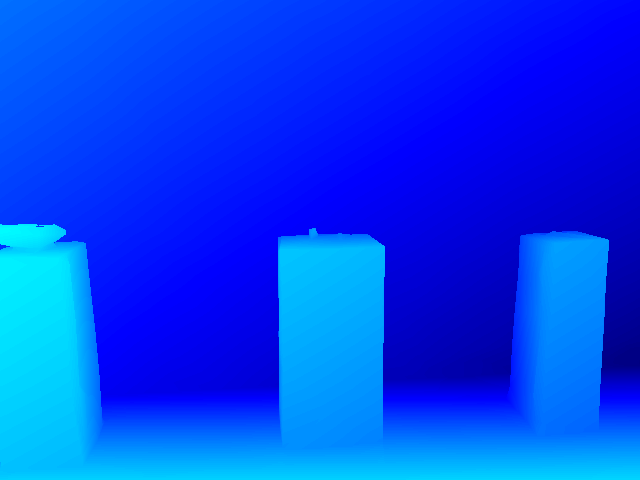}
        \end{subfigure}%
        \hfill
        \begin{subfigure}{0.161\linewidth}
            \centering
            \includegraphics[width=\linewidth]{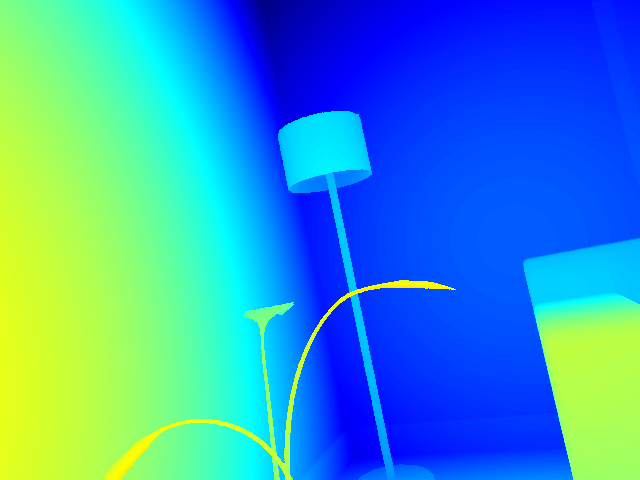}
        \end{subfigure}%
        \hfill
        \begin{subfigure}{0.161\linewidth}
            \centering
            \includegraphics[width=\linewidth]{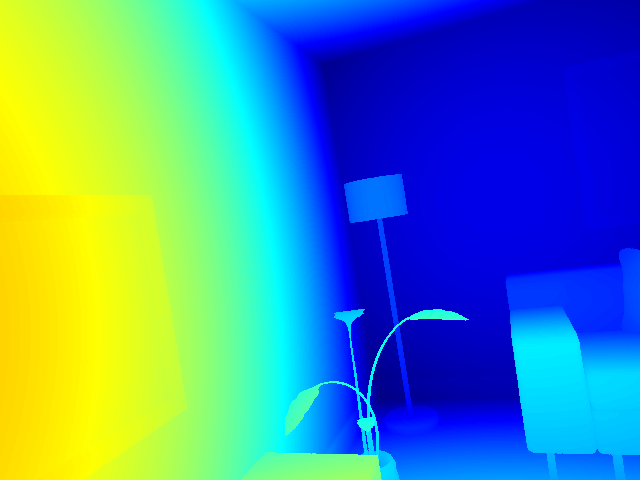}
        \end{subfigure}%
        \hfill
        \begin{subfigure}{0.161\linewidth}
            \centering
            \includegraphics[width=\linewidth]{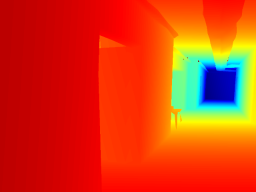}
        \end{subfigure}%
        \hfill
        \begin{subfigure}{0.161\linewidth}
            \centering
            \includegraphics[width=\linewidth]{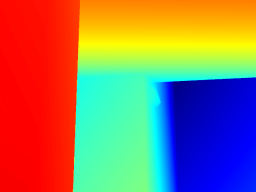}
        \end{subfigure}%
    \end{minipage}

\vspace{0.15em} 
    \begin{minipage}{0.05\linewidth}
        \centering
        \adjustbox{valign=c}{\rotatebox{90}{\fontsize{10pt}{10pt}\selectfont{PlaneAE}}}
    \end{minipage}%
    \hspace{-0.4em} 
    \begin{minipage}{0.9\linewidth} 
        \centering
        \begin{subfigure}{0.161\linewidth}
            \centering
            \includegraphics[width=\linewidth]{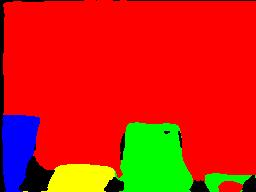}
        \end{subfigure}%
        \hfill 
        \begin{subfigure}{0.161\linewidth}
            \centering
            \includegraphics[width=\linewidth]{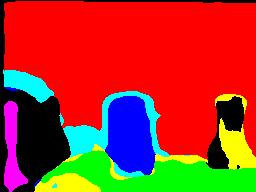}
        \end{subfigure}%
        \hfill
        \begin{subfigure}{0.161\linewidth}
            \centering
            \includegraphics[width=\linewidth]{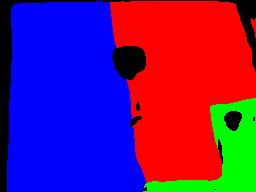}
        \end{subfigure}%
        \hfill
        \begin{subfigure}{0.161\linewidth}
            \centering
            \includegraphics[width=\linewidth]{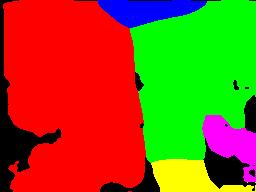}
        \end{subfigure}%
        \hfill
        \begin{subfigure}{0.161\linewidth}
            \centering
            \includegraphics[width=\linewidth]{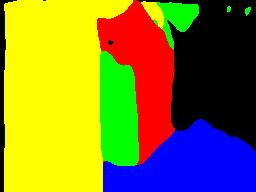}
        \end{subfigure}%
        \hfill
        \begin{subfigure}{0.161\linewidth}
            \centering
            \includegraphics[width=\linewidth]{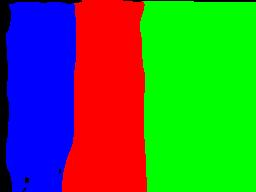}
        \end{subfigure}%
    \end{minipage}

\vspace{0.15em} 
    \begin{minipage}{0.05\linewidth}
        \centering
        \adjustbox{valign=c}{\rotatebox{90}{\fontsize{10pt}{10pt}\selectfont{PlaneTR}}}
    \end{minipage}%
    \hspace{-0.4em} 
    \begin{minipage}{0.9\linewidth} 
        \centering
        \begin{subfigure}{0.161\linewidth}
            \centering
            \includegraphics[width=\linewidth]{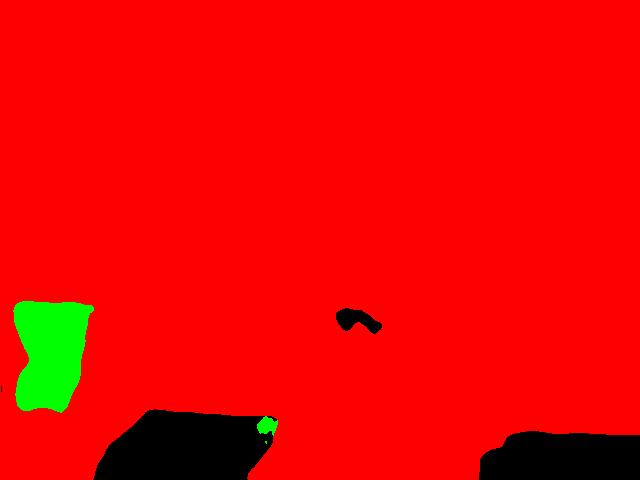}
        \end{subfigure}%
        \hfill 
        \begin{subfigure}{0.161\linewidth}
            \centering
            \includegraphics[width=\linewidth]{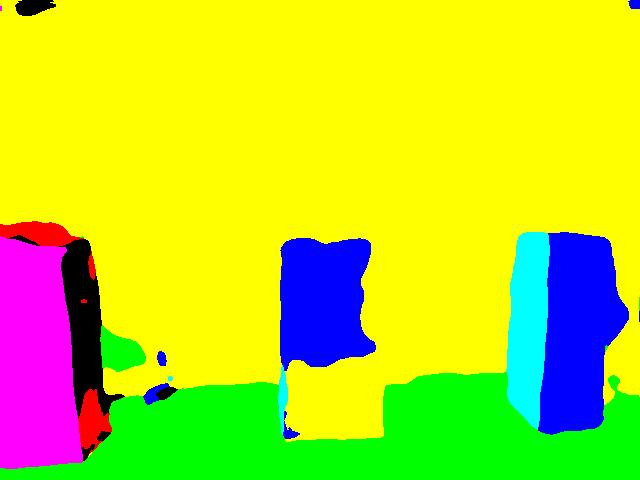}
        \end{subfigure}%
        \hfill
        \begin{subfigure}{0.161\linewidth}
            \centering
            \includegraphics[width=\linewidth]{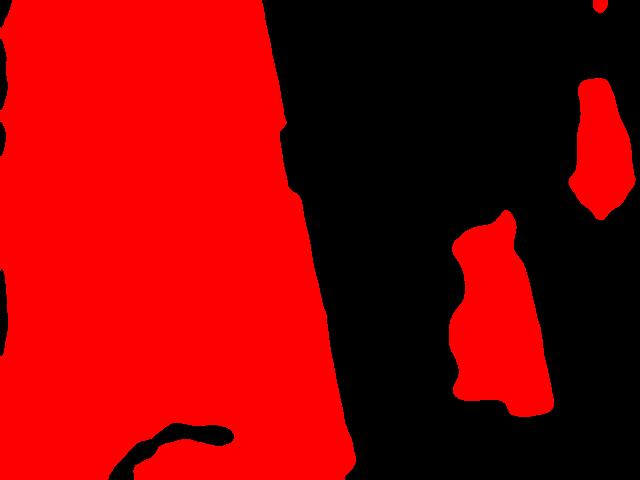}
        \end{subfigure}%
        \hfill
        \begin{subfigure}{0.161\linewidth}
            \centering
            \includegraphics[width=\linewidth]{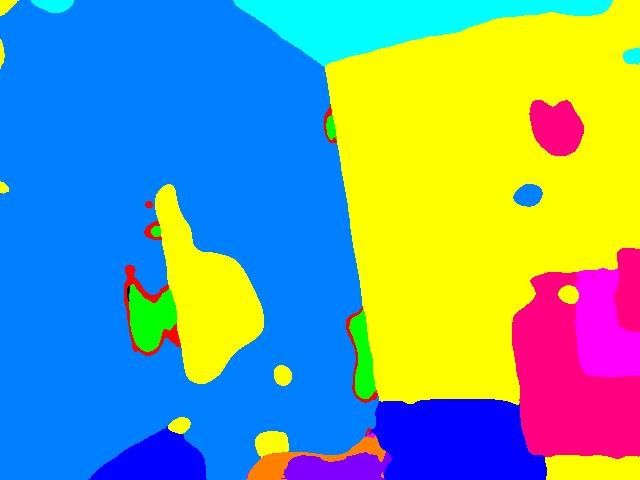}
        \end{subfigure}%
        \hfill
        \begin{subfigure}{0.161\linewidth}
            \centering
            \includegraphics[width=\linewidth]{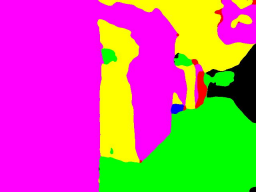}
        \end{subfigure}%
        \hfill
        \begin{subfigure}{0.161\linewidth}
            \centering
            \includegraphics[width=\linewidth]{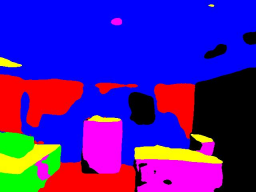}
        \end{subfigure}%
    \end{minipage}

\vspace{0.15em} 
    \begin{minipage}{0.05\linewidth}
        \centering
        \adjustbox{valign=c}{\rotatebox{90}{\fontsize{10pt}{10pt}\selectfont{X-PDNet}}}
    \end{minipage}%
    \hspace{-0.4em} 
    \begin{minipage}{0.9\linewidth} 
        \centering
        \begin{subfigure}{0.161\linewidth}
            \centering
            \includegraphics[width=\linewidth]{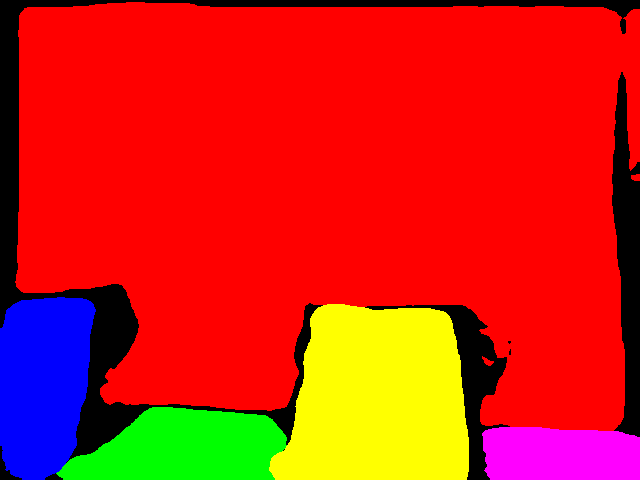}
        \end{subfigure}%
        \hfill 
        \begin{subfigure}{0.161\linewidth}
            \centering
            \includegraphics[width=\linewidth]{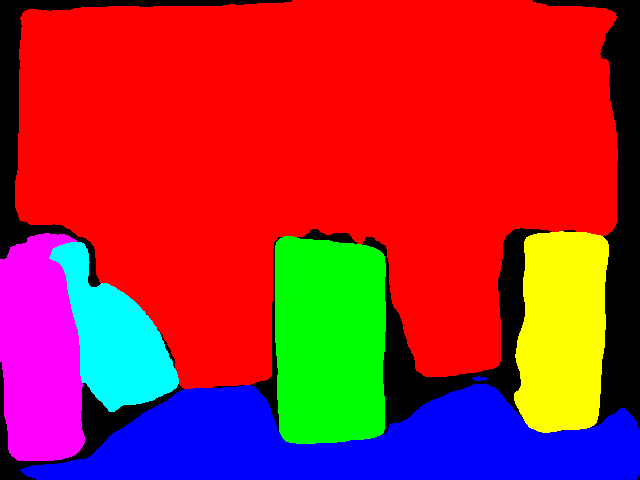}
        \end{subfigure}%
        \hfill
        \begin{subfigure}{0.161\linewidth}
            \centering
            \includegraphics[width=\linewidth]{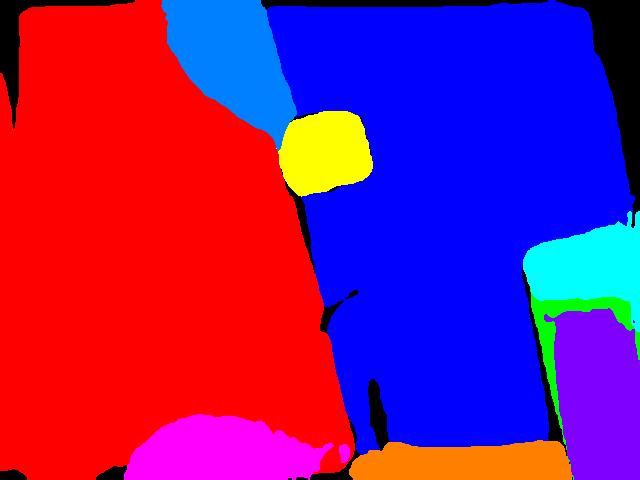}
        \end{subfigure}%
        \hfill
        \begin{subfigure}{0.161\linewidth}
            \centering
            \includegraphics[width=\linewidth]{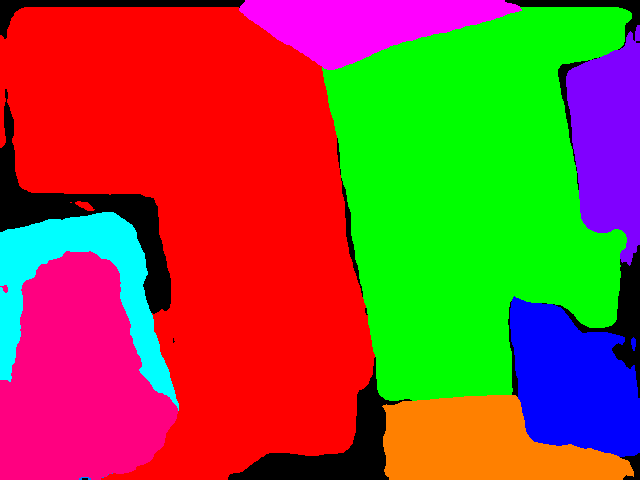}
        \end{subfigure}%
        \hfill
        \begin{subfigure}{0.161\linewidth}
            \centering
            \includegraphics[width=\linewidth]{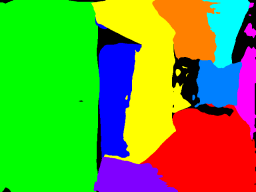}
        \end{subfigure}%
        \hfill
        \begin{subfigure}{0.161\linewidth}
            \centering
            \includegraphics[width=\linewidth]{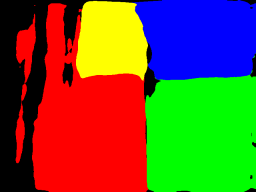}
        \end{subfigure}%
    \end{minipage}

\vspace{0.15em} 
    \begin{minipage}{0.05\linewidth}
        \centering
        \adjustbox{valign=c}{\rotatebox{90}{\fontsize{10pt}{10pt}\selectfont{Ours}}}
    \end{minipage}%
    \hspace{-0.4em} 
    \begin{minipage}{0.9\linewidth} 
        \centering
        \begin{subfigure}{0.161\linewidth}
            \centering
            \includegraphics[width=\linewidth]{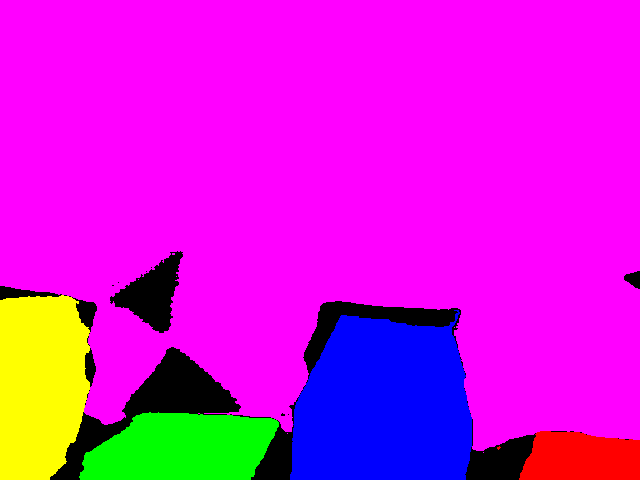}
        \end{subfigure}%
        \hfill 
        \begin{subfigure}{0.161\linewidth}
            \centering
            \includegraphics[width=\linewidth]{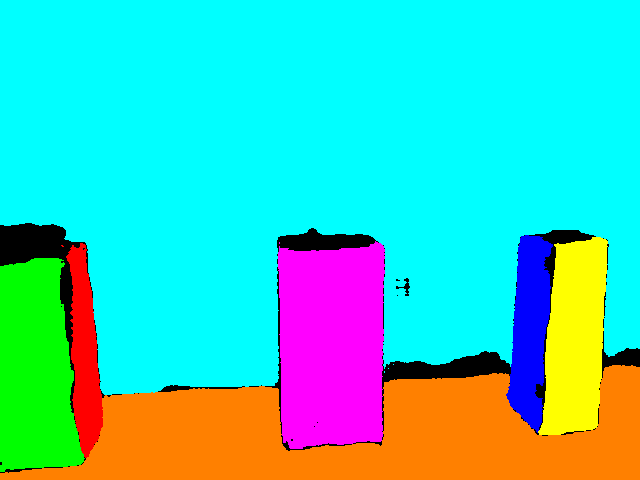}
        \end{subfigure}%
        \hfill
        \begin{subfigure}{0.161\linewidth}
            \centering
            \includegraphics[width=\linewidth]{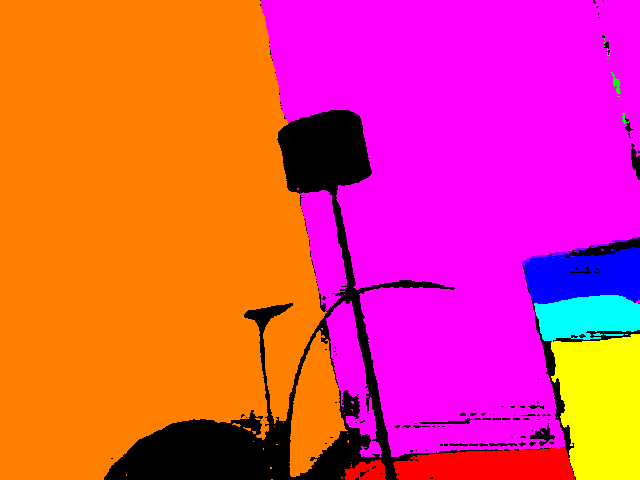}
        \end{subfigure}%
        \hfill
        \begin{subfigure}{0.161\linewidth}
            \centering
            \includegraphics[width=\linewidth]{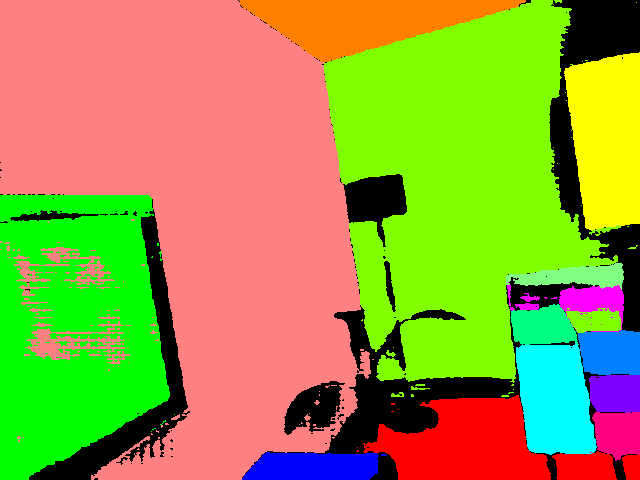}
        \end{subfigure}%
        \hfill
        \begin{subfigure}{0.161\linewidth}
            \centering
            \includegraphics[width=\linewidth]{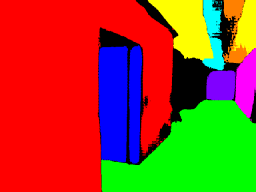}
        \end{subfigure}%
        \hfill
        \begin{subfigure}{0.161\linewidth}
            \centering
            \includegraphics[width=\linewidth]{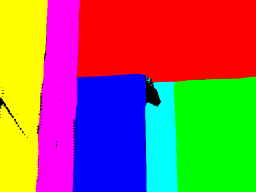}
        \end{subfigure}%
    \end{minipage}

\vspace{0.15em} 
    \begin{minipage}{0.05\linewidth}
        \centering
        \adjustbox{valign=c}{\rotatebox{90}{\fontsize{10pt}{10pt}\selectfont{GT}}}
    \end{minipage}%
    \hspace{-0.4em} 
    \begin{minipage}{0.9\linewidth} 
        \centering
        \begin{subfigure}{0.161\linewidth}
            \centering
            \includegraphics[width=\linewidth]{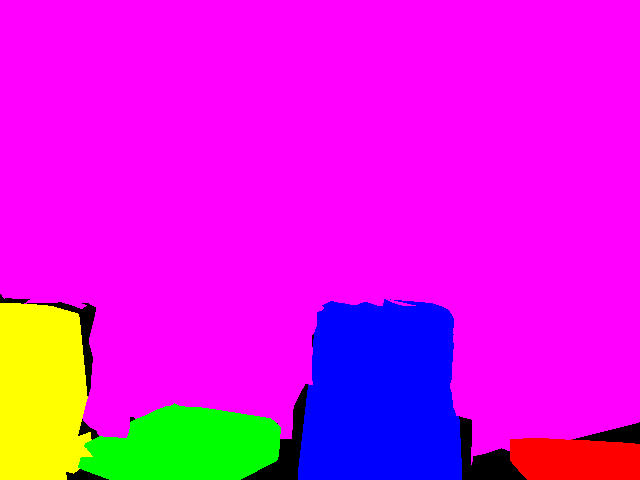}
        \end{subfigure}%
        \hfill 
        \begin{subfigure}{0.161\linewidth}
            \centering
            \includegraphics[width=\linewidth]{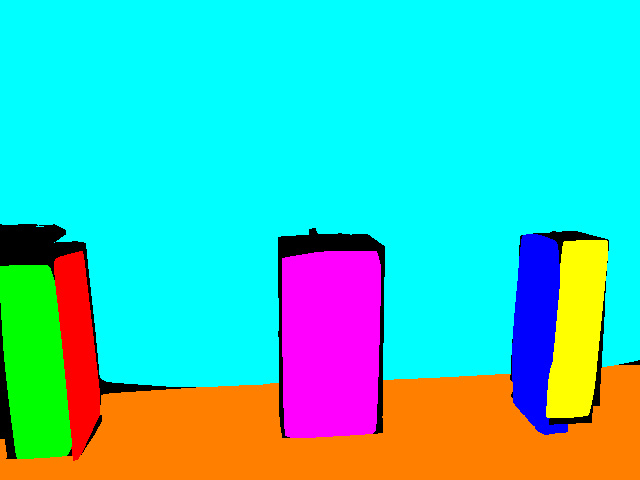}
        \end{subfigure}%
        \hfill
        \begin{subfigure}{0.161\linewidth}
            \centering
            \includegraphics[width=\linewidth]{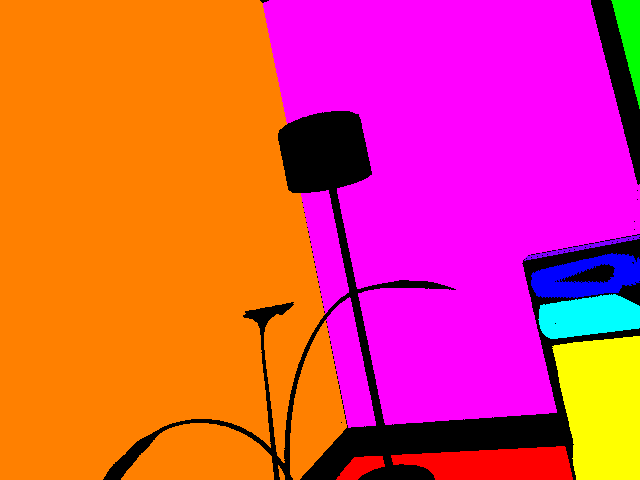}
        \end{subfigure}%
        \hfill
        \begin{subfigure}{0.161\linewidth}
            \centering
            \includegraphics[width=\linewidth]{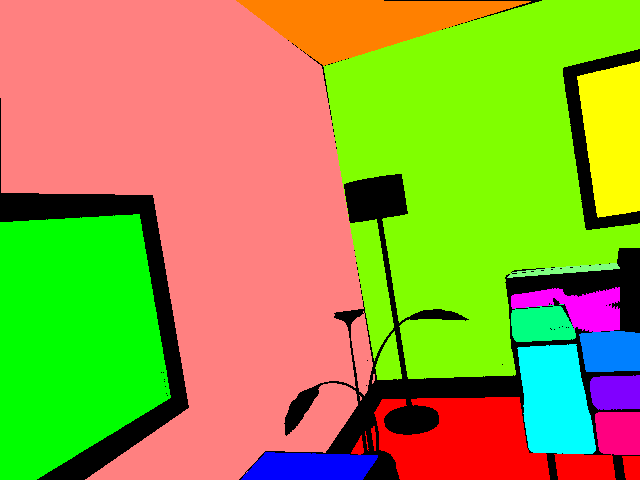}
        \end{subfigure}%
        \hfill
        \begin{subfigure}{0.161\linewidth}
            \centering
            \includegraphics[width=\linewidth]{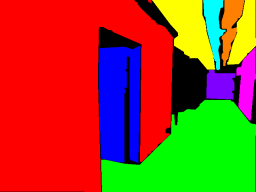}
        \end{subfigure}%
        \hfill
        \begin{subfigure}{0.161\linewidth}
            \centering
            \includegraphics[width=\linewidth]{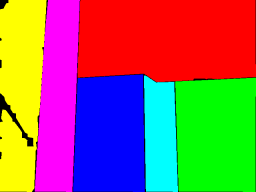}
        \end{subfigure}%
    \end{minipage}
    
    \vspace{0.4em} 
    \captionsetup{font=small}  
    \caption{Qualitative comparison results on unseen datasets. Columns 1-2, 3-4, and 5-6 show results on the Matterport3D, ICL-NUIM RGB-D, and 2D-3D-S datasets, respectively. GT refers to ground-truth.}
    \label{fig:Results_unseen}
\end{figure}

To evaluate our PlaneSAM's generalization capability, we tested all the compared methods, for which official implementations are available, on three unseen datasets: Matterport3D, ICL-NUIM RGB-D, and 2D-3D-S datasets. We trained them on only the ScanNet dataset, and tested them on these unseen datasets. 
The quantitative evaluation results are presented in Table~\ref{tab:Results_unseen}. 
On the Matterport3D and 2D-3D-S datasets, our PlaneSAM outperforms X-PDNet and PlaneTR---the second- and third-best performing methods---in terms of most of the metrics. 
On the ICL-NUIM RGB-D dataset, our PlaneSAM obtains better RI and SC measures, but a worse VOI measure, compared with X-PDNet and PlaneTR. 
This is because some depth images in the ICL-NUIM RGB-D dataset are very noisy, which significantly affects the predictive performance of our PlaneSAM; in contrast, because X-PDNet and PlaneTR are based only on RGB bands, they are not impacted by this disadvantage. 
Overall, our PlaneSAM exhibits highly competitive performance on all three datasets, outperforming the baselines in terms of generalization capabilities. 
Figure~\ref{fig:Results_unseen} presents visual comparison results of all test methods on the Matterport3D, ICL-NUIM RGB-D and 2D-3D-S datasets, with our PlaneSAM once again exhibiting the best performance.

\begin{table}[t!]
    \centering
    \small  
    \captionsetup{font=small}  
    \caption{Quantitative evaluation results on unseen datasets.}  
    \label{tab:Results_unseen}    
    \setlength{\tabcolsep}{4.5pt} 
    \begin{tabular}{>{\centering\arraybackslash}m{2.3cm} 
                    >{\centering\arraybackslash}m{4cm} >{\centering\arraybackslash}m{4cm} >{\centering\arraybackslash}m{4cm} 
                    >{\centering\arraybackslash}m{4cm} >{\centering\arraybackslash}m{4cm} >{\centering\arraybackslash}m{4cm} 
                    >{\centering\arraybackslash}m{4cm} >{\centering\arraybackslash}m{4cm} >{\centering\arraybackslash}m{4cm}} 
        \toprule
        \multirow{2}{*}{Method} & \multicolumn{3}{c}{Matterport3D} & \multicolumn{3}{c}{ICL-NUIM RGB-D} & \multicolumn{3}{c}{2D-3D-S} \\
        & \multicolumn{1}{c}{VOI\textdownarrow} & \multicolumn{1}{c}{RI\textuparrow} & \multicolumn{1}{c}{SC\textuparrow} 
        & \multicolumn{1}{c}{VOI\textdownarrow} & \multicolumn{1}{c}{RI\textuparrow} & \multicolumn{1}{c}{SC\textuparrow} 
        & \multicolumn{1}{c}{VOI\textdownarrow} & \multicolumn{1}{c}{RI\textuparrow} & \multicolumn{1}{c}{SC\textuparrow} \\
        \midrule
        PlaneAE   & \multicolumn{1}{c}{2.594} & \multicolumn{1}{c}{0.741} & \multicolumn{1}{c}{0.436} 
                  & \multicolumn{1}{c}{2.263} & \multicolumn{1}{c}{0.737} & \multicolumn{1}{c}{0.501} 
                  & \multicolumn{1}{c}{2.569} & \multicolumn{1}{c}{0.714} & \multicolumn{1}{c}{0.444} \\
        PlaneTR   & \multicolumn{1}{c}{2.654} & \multicolumn{1}{c}{0.758} & \multicolumn{1}{c}{0.441} 
                  & \multicolumn{1}{c}{\bfseries 1.809} & \multicolumn{1}{c}{0.784} & \multicolumn{1}{c}{0.603} 
                  & \multicolumn{1}{c}{2.456} & \multicolumn{1}{c}{0.724} & \multicolumn{1}{c}{0.479} \\
        X-PDNet   & \multicolumn{1}{c}{2.407} & \multicolumn{1}{c}{0.788} & \multicolumn{1}{c}{0.494} 
                  & \multicolumn{1}{c}{1.918} & \multicolumn{1}{c}{0.797} & \multicolumn{1}{c}{0.603} 
                  & \multicolumn{1}{c}{1.916} & \multicolumn{1}{c}{0.804} & \multicolumn{1}{c}{0.598} \\
        Ours      & \multicolumn{1}{c}{\bfseries 1.884} & \multicolumn{1}{c}{\bfseries 0.844} & \multicolumn{1}{c}{\bfseries 0.597} 
                  & \multicolumn{1}{c}{2.103} & \multicolumn{1}{c}{\bfseries 0.816} & \multicolumn{1}{c}{\bfseries 0.607} 
                  & \multicolumn{1}{c}{\bfseries 1.159} & \multicolumn{1}{c}{\bfseries 0.886} & \multicolumn{1}{c}{\bfseries 0.741} \\
        \bottomrule
    \end{tabular}
\end{table}

\subsection{Ablation experiments}  
\label{Sec:Ablation}
This section details the ablation experiments conducted to verify the effectiveness of our PlaneSAM's design and the pretraining strategy. 
We first conducted experiments to verify the validity of our dual-complexity backbone design. 
Then we conducted experiments to verify the contribution of pretraining (on the segment anything task) to our method. 
We also conducted experiments to determine the robustness of our PlaneSAM to noise in the prompt box. 
The efficiency of our dual-complexity backbone was also tested.

\textbf{Ablation experiments on not freezing the weights of EfficientSAM.} 
First, we conducted experiments to demonstrate that our dual-complexity backbone is superior to the dual-branch backbone structure that freezes the original branch's weights. 
If the weights of EfficientSAM are frozen, our dual-complexity backbone will become a structure similar to that of DPLNet~\cite{DPLNet2023}. 
However, as we argued earlier, we believe that our dual-complexity backbone can more effectively transfer a foundation model from the RGB domain to the RGB-D domain, especially when fine-tuning data is limited. 
The results in Tables \ref{tab:Ablation_backbone_ScanNet} and \ref{tab:Ablation_backbone_unseen}, where ``freeze SAM'' refers to freezing the weights of EfficientSAM, show that our dual-complexity backbone achieves better results across all datasets and evaluation metrics. 
This is because EfficientSAM was trained for the segment anything task, whereas our current task is plane instance segmentation, which differs significantly. Therefore, the weights of EfficientSAM require moderate fine-tuning: it needs to be fine-tuned, but not excessively, as that would damage the feature representations it learned from large-scale RGB data. 
Our dual-complexity backbone allows for this moderate fine-tuning. 
Additionally, since the branch that learns the D-band features is primarily a low-complexity CNN branch, it also helps to avoid overfitting caused by the small scale of the fine-tuning data. 
Therefore, the dual-complexity backbone proposed in this paper is well-suited for transferring the large RGB-domain model EfficientSAM to the task of RGB-D plane instance segmentation when the training dataset is limited.

\begin{table}[htbp]
    \centering
    \small  
    \captionsetup{font=small}  
    \caption{Ablation results for not freezing the weights of EfficientSAM on the ScanNet dataset.}   
    \label{tab:Ablation_backbone_ScanNet}
    \begin{tabular}{>{\centering\arraybackslash}m{3.6cm} >{\centering\arraybackslash}m{2.7cm} >{\centering\arraybackslash}m{2.7cm} >{\centering\arraybackslash}m{2.7cm}} 
        \toprule
        \multirow{2}{*}{{backbone}} & \multicolumn{3}{c}{{ScanNet}} \\
        & {VOI} $\downarrow$ & {RI} $\uparrow$ & {SC} $\uparrow$ \\
        \midrule
        freeze SAM & 0.735 & 0.918 & 0.826 \\
        Ours       & \textbf{0.550} & \textbf{0.941} & \textbf{0.873} \\
        \bottomrule
    \end{tabular}
\end{table}

\begin{table}[htbp]
    \centering
    \small  
    \captionsetup{font=small}  
    \caption{Ablation results for not freezing the weights of EfficientSAM on unseen datasets.}   
    \label{tab:Ablation_backbone_unseen}
    \setlength{\tabcolsep}{4.5pt} 
    \begin{tabular}{>{\centering\arraybackslash}m{2.3cm} 
                    >{\centering\arraybackslash}m{4cm} >{\centering\arraybackslash}m{4cm} >{\centering\arraybackslash}m{4cm} 
                    >{\centering\arraybackslash}m{4cm} >{\centering\arraybackslash}m{4cm} >{\centering\arraybackslash}m{4cm} 
                    >{\centering\arraybackslash}m{4cm} >{\centering\arraybackslash}m{4cm} >{\centering\arraybackslash}m{4cm}} 
        \toprule
        \multirow{2}{*}{backbone} & \multicolumn{3}{c}{Matterport3D} & \multicolumn{3}{c}{ICL-NUIM RGB-D} & \multicolumn{3}{c}{2D-3D-S} \\
        & \multicolumn{1}{c}{VOI$\downarrow$} & \multicolumn{1}{c}{RI$\uparrow$} & \multicolumn{1}{c}{SC$\uparrow$} 
        & \multicolumn{1}{c}{VOI$\downarrow$} & \multicolumn{1}{c}{RI$\uparrow$} & \multicolumn{1}{c}{SC$\uparrow$} 
        & \multicolumn{1}{c}{VOI$\downarrow$} & \multicolumn{1}{c}{RI$\uparrow$} & \multicolumn{1}{c}{SC$\uparrow$} \\
        \midrule
        freeze SAM & \multicolumn{1}{c}{2.091} & \multicolumn{1}{c}{0.798} & \multicolumn{1}{c}{0.544}
                  & \multicolumn{1}{c}{2.989} & \multicolumn{1}{c}{0.745} & \multicolumn{1}{c}{0.507}
                  & \multicolumn{1}{c}{1.496} & \multicolumn{1}{c}{0.828} & \multicolumn{1}{c}{0.651} \\
        Ours      & \multicolumn{1}{c}{\textbf{1.884}} & \multicolumn{1}{c}{\textbf{0.844}} & \multicolumn{1}{c}{\textbf{0.597}}
                  & \multicolumn{1}{c}{\textbf{2.103}} & \multicolumn{1}{c}{\textbf{0.816}} & \multicolumn{1}{c}{\textbf{0.607}}
                  & \multicolumn{1}{c}{\textbf{1.159}} & \multicolumn{1}{c}{\textbf{0.886}} & \multicolumn{1}{c}{\textbf{0.741}} \\
        \bottomrule
    \end{tabular}
\end{table}

\textbf{Ablation experiments on the low-complexity CNN branch.}
Then, we validated the contribution of the low-complexity CNN branch (which is connected in parallel to EfficientSAM) to our PlaneSAM. 
As shown in Table~\ref{tab:ablation_CNNBranch_pretraining_ScanNet}, the CNN branch significantly improves the performance of our PlaneSAM, with the VOI value decreasing by approximately 0.7 and RI and SC values increasing by approximately 0.9 and 0.17, respectively. 
It can be said that without mainly deploying the CNN branch to learn the D-band features---i.e., only fine-tuning EfficientSAM~\cite{Xiong_CVPR2024} on the ScanNet dataset---it would be difficult to achieve high-quality plane instance segmentation. 
The main reason is that the original EfficientSAM focuses on spectral features, whereas plane instance segmentation relies heavily on geometric features, which means D-band features need to be learned.
However, in the context where the backbone network of EfficientSAM is complex and D-band training data is limited, the absence of the low-complexity CNN branch fails to effectively mitigate overfitting. 
The ablation experiments in Table~\ref{tab:ablation_CNNBranch_pretraining_ScanNet} also show that the CNN branch can fully integrate the features of the RGB and depth bands.

\begin{table}[htbp]
    \centering
    \small  
    \captionsetup{font=small}  
    \caption{Ablation results for the low-complexity CNN branch and pretraining on the ScanNet dataset.}   
    \label{tab:ablation_CNNBranch_pretraining_ScanNet}
    \begin{tabular}{>{\centering\arraybackslash}m{2.5cm} >{\centering\arraybackslash}m{2.5cm} >{\centering\arraybackslash}m{2.5cm} >{\centering\arraybackslash}m{2.5cm} >{\centering\arraybackslash}m{2.5cm}} 
        \toprule
        \multicolumn{2}{c}{Setting} & \multicolumn{1}{c}{\multirow{2}{*}{VOI $\downarrow$}} & \multicolumn{1}{c}{\multirow{2}{*}{RI $\uparrow$}} & \multicolumn{1}{c}{\multirow{2}{*}{SC $\uparrow$}} \\
        CNN branch & pretraining & & & \\
        \midrule
         &  & 1.295 & 0.847 & 0.700 \\
        \checkmark&  & 0.572 & 0.932 & 0.868 \\
         & \checkmark & 1.247 & 0.873 & 0.739 \\
        \checkmark&\checkmark & \textbf{0.550} & \textbf{0.941} & \textbf{0.873} \\
        \bottomrule
    \end{tabular}

    \vspace{5pt} 

    \parbox{\textwidth}{\#The first case (third row) indicates that the low-complexity CNN branch is not added, and there is no pretraining, 
        which means that the model is obtained by directly fine-tuning EfficientSAM. The third case (fifth row) 
        indicates that the CNN branch was not added, but EfficientSAM is pretrained on the ScanNet dataset.}

\end{table}

\textbf{Ablation experiments on pretraining.}
We also validate the contribution of pretraining to our PlaneSAM. Table~\ref{tab:ablation_CNNBranch_pretraining_ScanNet} shows the ablation experiment results on the ScanNet dataset, the training set of which has been used to fine-tune our PlaneSAM. 
As shown in Table~\ref{tab:ablation_CNNBranch_pretraining_ScanNet}, pretraining contributes to an improvement in plane instance segmentation across all evaluation metrics. 
This improvement is attributed to the ability of our PlaneSAM to learn more comprehensive feature representations of RGB-D data after pretraining on additional RGB-D images. 
Additionally, the plane instance segmentation task shares certain similarities with the segment anything task, facilitating a smoother transfer to plane instance segmentation.

Table~\ref{tab:ablation_pretraining_unseen} shows ablation experiment results on unseen datasets, which were not used for fine-tuning the model for plane instance segmentation. 
Specifically, pretraining in Table~\ref{tab:ablation_pretraining_unseen} means that our PlaneSAM was pretrained for the segment anything task using 100,000 images from the ScanNet\_25k~\cite{Dai_CVPR2017}, SUN RGB-D~\cite{Song_CVPR2015} and 2D-3D-S~\cite{2D3DS_2017} datasets, but only fine-tuned using the ScanNet dataset for plane instance segmentation. 
From Table~\ref{tab:ablation_pretraining_unseen}, we can see that the pretrained model outperforms the non-pretrained model across all three evaluation metrics, particularly on the ICL-NUIM RGB-D dataset. 
This demonstrates that after relevant pretraining, EfficientSAM exhibits enhanced adaptability to the RGB-D data domain. 
Besides, this also indicates that Faster R-CNN has good generalization capabilities in the plane detection task, often providing effective box prompts on unseen data.

\begin{table}[htbp]
    \centering
    \small  
    \captionsetup{font=small}  
    \caption{Ablation results for pretraining on unseen datasets.}    
    \label{tab:ablation_pretraining_unseen}    
    \setlength{\tabcolsep}{4.5pt} 
    \begin{tabular}{>{\centering\arraybackslash}m{2.3cm} 
                    >{\centering\arraybackslash}m{7cm} >{\centering\arraybackslash}m{7cm} >{\centering\arraybackslash}m{7cm} 
                    >{\centering\arraybackslash}m{7cm} >{\centering\arraybackslash}m{7cm} >{\centering\arraybackslash}m{7cm} 
                    >{\centering\arraybackslash}m{7cm} >{\centering\arraybackslash}m{7cm} >{\centering\arraybackslash}m{7cm}} 
        \toprule
        \multirow{2}{*}{Method} & \multicolumn{3}{c}{Matterport3D} & \multicolumn{3}{c}{ICL-NUIM RGB-D} & \multicolumn{3}{c}{2D-3D-S} \\
        & \multicolumn{1}{c}{VOI$\downarrow$} & \multicolumn{1}{c}{RI$\uparrow$} & \multicolumn{1}{c}{SC$\uparrow$} 
        & \multicolumn{1}{c}{VOI$\downarrow$} & \multicolumn{1}{c}{RI$\uparrow$} & \multicolumn{1}{c}{SC$\uparrow$} 
        & \multicolumn{1}{c}{VOI$\downarrow$} & \multicolumn{1}{c}{RI$\uparrow$} & \multicolumn{1}{c}{SC$\uparrow$} \\
        \midrule
        no pretraining & \multicolumn{1}{c}{2.029} & \multicolumn{1}{c}{0.820} & \multicolumn{1}{c}{0.565} 
                       & \multicolumn{1}{c}{3.254} & \multicolumn{1}{c}{0.726} & \multicolumn{1}{c}{0.460}
                       & \multicolumn{1}{c}{1.322} & \multicolumn{1}{c}{0.854} & \multicolumn{1}{c}{0.703} \\
        pretraining    & \multicolumn{1}{c}{\textbf{1.884}} & \multicolumn{1}{c}{\textbf{0.844}} & \multicolumn{1}{c}{\textbf{0.597}} 
                       & \multicolumn{1}{c}{\textbf{2.103}} & \multicolumn{1}{c}{\textbf{0.816}} & \multicolumn{1}{c}{\textbf{0.607}}
                       & \multicolumn{1}{c}{\textbf{1.159}} & \multicolumn{1}{c}{\textbf{0.886}} & \multicolumn{1}{c}{\textbf{0.741}} \\
        \bottomrule
    \end{tabular}
\end{table}

\textbf{Ablation experiments on box prompt noise.}
We also performed ablation experiments on the quality of box prompts, with detailed results listed in Table~\ref{tab:Ablation_box_prompt_quality}, where n\% indicates the addition of random noise, ranging from 0 to n\% of the bounding box length, to the four vertex coordinates of the ground-truth bounding box. 
Using the no-noise condition as the baseline (second column in the table), we applied 0-n\% noise to the box prompts. 
As shown in Table~\ref{tab:Ablation_box_prompt_quality}, when using the ground-truth bounding boxes as prompts, the segmentation masks of our method closely approximate the ground-truth masks, and the introduction of 0-10\% noise results in a minimal accuracy reduction of less than 1\%. 
Even with 0-20\% noise, our method maintains a very high accuracy. Although the accuracy significantly decreases with 0-30\% noise, the segmentation results remain competitive. 
The above results demonstrate that our PlaneSAM can achieve high-quality plane instance segmentation even when the accuracy of the box prompts is not high. 
This capability enables the proposed method to achieve excellent results in complex scenes.

\begin{table}[htbp]
    \centering
    \small  
    \captionsetup{font=small}  
    \caption{Ablation results for box prompt quality.}   
    \label{tab:Ablation_box_prompt_quality}
    \begin{tabular}{>{\centering\arraybackslash}m{2.5cm} >{\centering\arraybackslash}m{2.5cm} >{\centering\arraybackslash}m{2.5cm} >{\centering\arraybackslash}m{2.5cm} >{\centering\arraybackslash}m{2.5cm}} 
        \toprule
        \multirow{2}{*}{Metric} & \multicolumn{4}{c}{Noise} \\
        & 0\% & 10\% & 20\% & 30\% \\
        \midrule
        VOI$\downarrow$ & \textbf{0.227} & 0.251 & 0.481 & 0.857 \\
        RI$\uparrow$    & \textbf{0.982} & 0.978 & 0.945 & 0.886 \\
        SC$\uparrow$    & \textbf{0.961} & 0.955 & 0.892 & 0.788 \\
        \bottomrule
    \end{tabular}
\end{table}

\textbf{Ablation experiments on processing efficiency.}
We also compared the training and testing speeds of our PlaneSAM with those of the model obtained by directly fine-tuning EfficientSAM for RGB-D plane instance segmentation. 
For training speed, we trained both models for 10 epochs and calculated the average processing time. 
We found that directly fine-tuning EfficientSAM required 46.28 minutes per epoch, while our PlaneSAM required 51.13 minutes per epoch, indicating the training speed of our PlaneSAM is merely 9.49\% slower than that of directly fine-tuning EfficientSAM. 
To evaluate testing speed, both models processed the same set of 1,000 images and the total processing time was recorded for each model. 
We found this processing time was 261 seconds for the fine-tuned EfficientSAM and 291 seconds for our PlaneSAM, with the testing speed of our model being only 10.30\% slower than that of the baseline. 
Thus, our PlaneSAM increases the computational overhead by only approximately 10\% compared to the model obtained by directly fine-tuning EfficientSAM.

\subsection{Discussion}  
\label{Sec:Discussion}
As for the comparison methods, both PlaneTR~\cite{Tan_ICCV2021} and PlaneAC~\cite{Zhang_PR2024} first extract line segments from the RGB bands using existing robust line extraction techniques, and then utilize these segments as geometric clues to aid in segmenting plane instances. 
Both methods perform well in the experiments on the ScanNet dataset (see Figure~\ref{fig:Results_Scannet} and Table~\ref{tab:Results_ScanNet}), indicating that relevant geometric clues are indeed useful for plane instance segmentation. 
However, since these geometric clues are extracted from RGB bands, both PlaneTR and PlaneAC are unable to fully leverage the geometric information in RGB-D data for effective plane segmentation.

PlaneTR performs well across all four datasets, while X-PDNet~\cite{Cao_BMVC2023} ranks second on the three unseen datasets (see Figure~\ref{fig:Results_unseen} and Table~\ref{tab:Results_unseen}). 
Since both methods simultaneously handle plane instance segmentation and monocular depth estimation, their strong performance in our experiments suggests that multi-task learning that integrates plane instance segmentation with monocular depth estimation greatly enhances the training of plane instance segmentation networks. 
However, our PlaneSAM, which uses a single-task learning approach, achieves even better results, demonstrating that effectively utilizing depth band information is highly advantageous for plane segmentation from RGB-D data.

If we observe the test methods across all four datasets, we find that although X-PDNet does not perform as well as other algorithms on the ScanNet dataset, its performance is quite stable across all four datasets: the evaluation metrics show little fluctuation, with the RI metric consistently around 0.800. Considering that X-PDNet uses depth information estimated from the RGB bands for plane segmentation, we conclude that its stable performance across the four datasets further indicates that using depth information is highly beneficial for RGB-D data plane segmentation. In contrast to X-PDNet, which uses depth information estimated from the RGB bands, our PlaneSAM directly utilizes the depth band of the RGB-D data and is more powerful.

From the above ablation experiments on pretraining, we can conclude that our pretraining strategy improves the model's generalization performance (see Tables \ref{tab:ablation_CNNBranch_pretraining_ScanNet} and \ref{tab:ablation_pretraining_unseen}), especially on unseen datasets (see Table~\ref{tab:ablation_pretraining_unseen}). 
These successful pretraining results suggest that imperfect pseudo-labels from algorithms for related tasks using the same type of data can still benefit model training. 
Compared with traditional pretraining that relies on manual fine annotations~\cite{He_TPAMI2020, DINO2023, DETR2020}, our pretraining strategy is much more cost-effective and is therefore better suited for pretraining with large-scale data. 
Therefore, when pretraining networks for other tasks where dataset annotation is challenging, in addition to using costly manually annotated datasets, we might also leverage other networks that process the same type of data but perform different tasks (preferably those with relatively good performance) to generate pseudo-labels, and use those pseudo-labels for pretraining. 
For example, annotating training sets for 3D point cloud instance segmentation is quite challenging, while annotating training sets for 3D point cloud classification is relatively easier. 
Therefore, when training 3D point cloud instance segmentation networks, besides using existing manually annotated 3D point cloud classification datasets for pretraining, we can also use pseudo-labels generated by existing 3D point cloud classification networks on large-scale 3D point cloud data for pretraining.

\section{Conclusion}
\label{sec:Conclusion}
In this study, we developed PlaneSAM, which, to the best of our knowledge, is the first deep learning model that utilizes all four bands of RGB-D data for plane instance segmentation. It is an extension of EfficientSAM designed to segment plane instances from RGB-D data in a multimodal manner, which is realized by adopting a dual-complexity network structure, modifying the loss function of EfficientSAM, and pretraining the model on the segment anything task using additional RGB-D datasets. We conducted comparative experiments with prior SOTA methods and necessary ablation experiments, all of which validate the effectiveness of our PlaneSAM. However, the proposed PlaneSAM has certain limitations, such as being susceptible to noise in depth images. In addition, it relies on the accuracy of the plane detection stage. Therefore, enhancing robustness to noise in depth images and improving predicted bounding box accuracy will be important research directions for our future work.

\section*{Acknowledgment}
This work was supported by the Open Research Fund Program of LIESMARS (No. 22S04), the Knowledge Innovation Program of Wuhan-Shuguang Project (No. 2023020201020414), the National Natural Science Foundation of China (No. U22A20568, No. 42071451, No. 42101446), the National Key RESEARCH and Development Program (No. 2022YFB3904101), the Key RESEARCH and Development Program of Hubei Province (No. 2023BAB146, No. 2021BAA188), the Research Program of State Grid Corporation of China (5500-202316189A-1-1-ZN), and the Fundamental Research Funds for the Central Universities, LIESMARS Special Research Funding.

We would also like to thank the Wuhan Supercomputing Center and the Supercomputing Center of Wuhan University for providing computational resources.

 \bibliographystyle{elsarticle-num} 
 \bibliography{refs}




\end{CJK}
\end{document}